\documentclass{article}

    \PassOptionsToPackage{numbers}{natbib}

    \usepackage[final]{neurips_2022}

\usepackage{enumitem}
\usepackage[utf8]{inputenc} 
\usepackage[T1]{fontenc}    
\usepackage{hyperref}       
\usepackage{url}            
\usepackage{booktabs}       
\usepackage{amsfonts}       
\usepackage{nicefrac}       
\usepackage{microtype}      
\usepackage[dvipsnames]{xcolor}       
\usepackage{wrapfig}

\newcommand{\actionseq}{{\bf a}}

\usepackage{amsmath,amsfonts,bm}

\def\eqref#1{equation~\ref{#1}}

\def\1{\bm{1}}

\DeclareMathAlphabet{\mathsfit}{\encodingdefault}{\sfdefault}{m}{sl}
\SetMathAlphabet{\mathsfit}{bold}{\encodingdefault}{\sfdefault}{bx}{n}

\DeclareMathOperator*{\argmax}{arg\,max}
\DeclareMathOperator*{\argmin}{arg\,min}

\usepackage{algorithm,algorithmicx,algpseudocode}
\usepackage{graphicx}
\usepackage{algorithm}
\usepackage{xr}
\usepackage[scientific-notation=true]{siunitx}

\title{ASPiRe:\\ Adaptive Skill Priors for  Reinforcement Learning}

\author{%
  Mengda Xu \textsuperscript{\rm 1, 2}, 
  Manuela Veloso \textsuperscript{\rm 2,3}, 
  Shuran Song \textsuperscript{\rm 1} \\
  \textsuperscript{\rm 1} Department of Computer Science, Columbia University \\
  \textsuperscript{\rm 2} J.P. Morgan AI Research  
  \textsuperscript{\rm 3} School of Computer Science, Carnegie Mellon University (emeritus)\\
\url{https://aspire.cs.columbia.edu/}
}

\begin{document}

\maketitle

\begin{abstract}
We introduce ASPiRe (Adaptive Skill Prior for RL), a new approach that leverages prior experience to accelerate reinforcement learning. Unlike existing methods that learn a single skill prior from a large and diverse dataset, our framework learns a library of different distinction skill priors (i.e., behavior priors) from a collection of specialized datasets, and learns how to combine them to solve a new task. This formulation allows the algorithm to acquire a set of specialized skill priors that are more reusable for downstream tasks; however, it also brings up additional challenges of how to effectively combine these unstructured sets of skill priors to form a new prior for new tasks. Specifically, it requires the agent not only to identify which skill prior(s) to use but also how to combine them (either sequentially or concurrently) to form a new prior. To achieve this goal, ASPiRe includes Adaptive Weight Module (AWM) that learns to infer an adaptive weight assignment between different skill priors and uses them to guide policy learning for downstream tasks via weighted Kullback-Leibler divergences. Our experiments demonstrate that ASPiRe can significantly accelerate the learning of new downstream tasks in the presence of multiple priors and show improvement on competitive baselines.          
\end{abstract}

\section{Introduction}
Transferring prior experience to new tasks is central to an agent's adaptability. In this work, we aim to accelerate online reinforcement learning by leveraging prior experience from large offline data. Previous works \citep{ajay2020opal,pertsch2020spirl,singh2020parrot} have proposed to extract temporally extended actions into \textit{skills} from offline datasets and learn a single behavior prior over the skill space, which we refer to as \textit{skill prior}. The skill prior is then used to guide the learning for a new downstream task by constraining the learned policy to stay close to the prior distribution. However, regularizing with a single skill prior might be insufficient for more complex downstream tasks. We conclude two major limitations for using a single skill prior: (1) The datasets could be generated by multiple policies for diverse tasks. As a consequence, the action distribution from behavior trajectories might be multi-modal, and the resulting learned prior might learn an "average" behavior \citep{DBLP:journals/corr/abs-2005-01643} (2) Complex tasks not only require reusing previous skills but also composing them. Consider a task that requires the agent to traverse a maze with obstacles inside. The agent needs to compose navigation skills and avoid skills concurrently when the obstacle is observed (See Fig. \ref{fig:overview}) and only activate navigation skills when it is not. Guided with only navigation skill prior or avoid skill prior might lead to inefficient learning. 

This observation suggests guiding the learning with multiple skill priors, with each prior specialized in one task. During the learning of new downstream tasks, the most relevant priors should be selected to regularize the policy based on the task and specific states. 
Specifically, it requires the agent to: 
(1) \textbf{Identify which skill prior(s) to use.} Since not all skill priors are relevant to the new task at hand, imposing irrelevant skill priors to regularize can mislead the learning process. 
(2) \textbf{Learn how to combine multiple skill priors.} Many complex tasks require the agent to identify and combine multiple priors to solve them effectively. The integration of skill priors can be either \textbf{sequential} (active one prior at a time) or \textbf{concurrent} (active multiple priors at the same time), which need to be determined by the agent automatically.

To address the above challenges, we propose \textbf{ASPiRe}, a new method to accelerate the learning of new downstream tasks by guiding the policy with an adaptive skill prior. To implement this idea, we first extract the multiple primitive skill priors from the separated and labeled offline data. Each primitive skill prior is specialized in solving one specific task (e.g., navigation prior and avoid prior). For a new downstream task, the policy is trained with a new task reward and regularized by the adaptive skill prior. The adaptive skill prior can be one of the primitive skill priors or constructed by composing primitive skill priors.

Our method includes an Adaptive Weight Module (AWM) that infers an adaptive weight assignment over learned primitive skill priors and uses these inferred weights to guide policy learning through weighted Kullback-Leibler (KL) divergence. The weight applied to the KL divergence between the learned policy and a given primitive skill prior determines its impact on the learned policy. This implicitly composes multiple primitive skill priors into a \textit{composite skill prior} and guides the policy learning. Our formulation allows the algorithm to combine the primitive skill priors both sequentially and concurrently as well as adaptively adjust them during the online learning phase.

We evaluate our method in challenging tasks with sparse rewards. We demonstrate that our method can accelerate the learning of new downstream tasks in the presence of multiple priors and show improvement over competitive baselines on both learning efficiency and task success rate.

\section{Related work}
\textbf{Offline reinforcement learning.} Offline reinforcement learning \citep{DBLP:journals/corr/abs-2005-01643} exploits an existing offline dataset and learns the best policy from it without interacting with environments. This is valuable when interacting with the environment for a task is expensive. Many algorithms have been developed, both in model-free \citep{Kumar2019StabilizingOQ,DBLP:journals/corr/abs-1911-11361,DBLP:journals/corr/abs-1904-08473,Fujimoto2019OffPolicyDR,kumar2020conservative,siegel2019keep,wang2020critic,DBLP:journals/corr/abs-1907-04543,DBLP:journals/corr/abs-1907-00456} and model-based \citep{kidambi2020morel,yu2020mopo,argenson2020model,matsushima2020deployment,cang2021behavioral} settings. 

Offline RL has also demonstrated its ability to accelerate online learning \cite{pertsch2020spirl,singh2020parrot,DBLP:journals/corr/abs-2006-09359}. While most of the previous works only consider a behavior prior for a single task, the policy is guided by the behavior prior to achieve efficient exploration. \citet{rao2021learning} proposes to use a mixture latent variable model to compose multiple behavior priors, but this framework only activates one behavior prior at a time. How to leverage multiple behavior priors during online learning both sequentially and concurrently remains an open question. Our method provides a solution to adaptively select a behavior prior (i.e., skill prior) or compose a new prior to guide the policy learning.

\textbf{Reusing and Composing skills.}
It is crucial for intelligent agents to reuse and compose learned skills to solve new tasks. A wide range of works have been proposed to solve this longstanding problem \citep{10.1145/1160633.1160762,article}. One way to compose and reuse skills is using value functions \citep{DBLP:journals/corr/abs-1803-06773,Hunt2019ComposingEP,pmlr-v97-van-niekerk19a} or cumulants \citep{barreto2017successor,barreto2018transfer,borsa2018universal,Barreto2019TheOK}. Numerous works also explore the hierarchical structure in which a high-level policy integrates a collection of low-level controllers (primitives) \citep{bacon2016optioncritic,hausknecht2016deep,frans2018meta,merel2018hierarchical,DBLP:journals/corr/abs-1906-11228} and each of low-level controllers contains a distinct skill. However, most of the hierarchical methods only sequentially activate the primitives. The other line of work for reusing and composing skills is to operate on a latent space, which can be mapped to action space \citep{heess2016learning,DBLP:journals/corr/FlorensaDA17,merel2018neural,hausman2018learning,eysenbach2018diversity,Haarnoja2018LatentSP}. Among those works, several approaches employ the idea of mixture-of-experts \citep{6797059} by explicitly modeling a library of primitive skills and learning a weight gate function to compose the skills \citep{Qureshi2020ComposingTP,tseng2021toward,Qureshi2020ComposingTP,ren2021probabilistic}. These works show that this additional structure is beneficial for transferring skills to complex tasks. Our work can also be viewed through the mixture-of-experts lens, i.e., solve downstream tasks by composing a library of primitives. However, our work treats the composition as a regularization to guide the policy learning, instead of executing the composition as a policy directly. 

\section{Approach}

\begin{figure}
\begin{center}
\includegraphics[width=\textwidth]{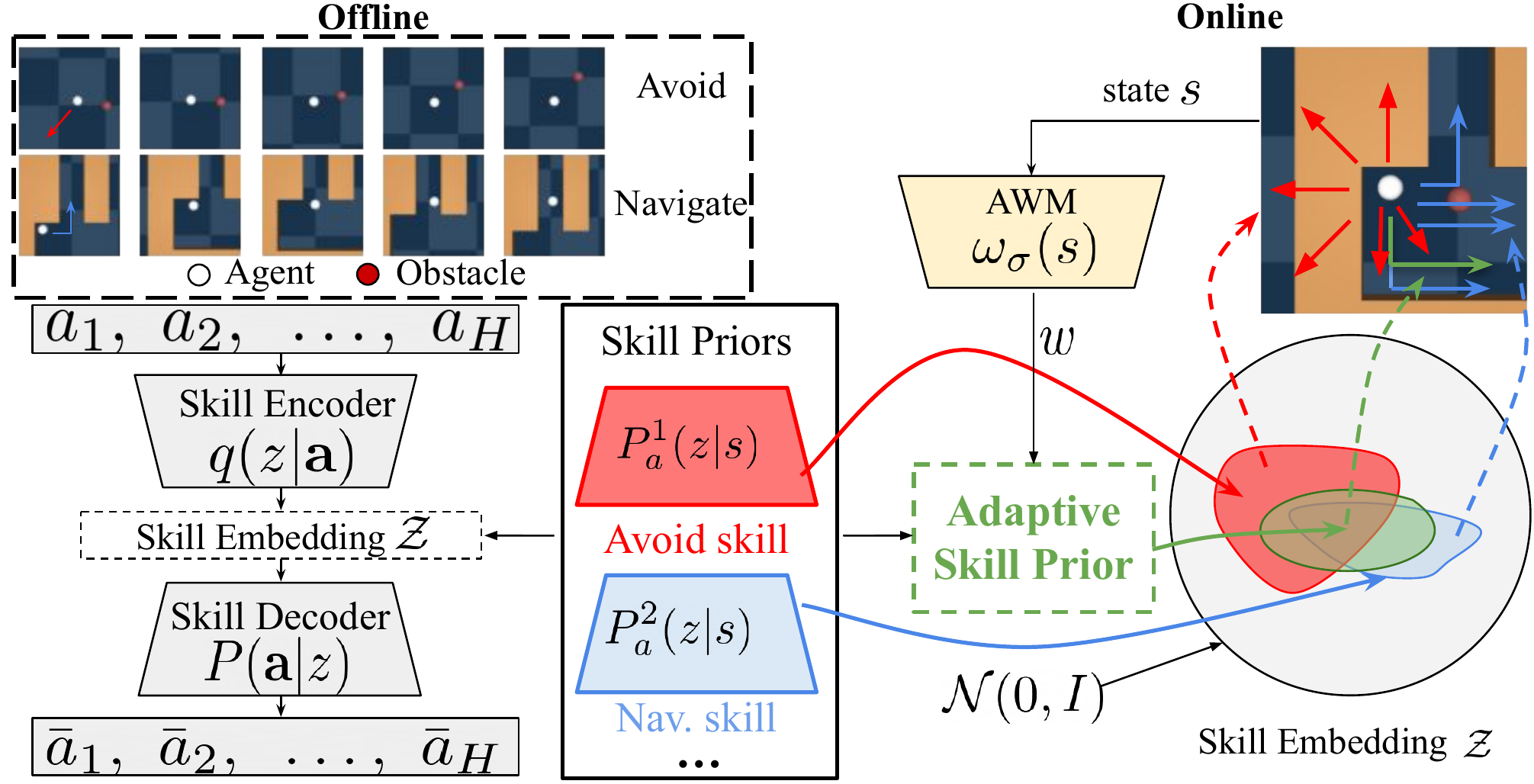} 
\caption{\textbf{ASPiRe overview.} In offline, we learn a shared skill embedding space and a collection of distinct primitive skill priors, e.g. , avoid and navigation primitives are learned. The skill encoder embeds an action sequence $\actionseq$, which is sampled from aggregated dataset, into a latent vector in embedding $\mathcal{Z}$ and the action sequence get reconstructed by the skill decoder. The primitive skill prior $P_a^i(z|s)$ imitates the state-skill (in embedding space) pairs in dataset $\mathcal{D}_i$. In online stage, ASPiRe composes multiple primitive skill priors via weighted KL divergence, where weighs are inferred by Adaptive Weight Module (AWM).}
\label{fig:overview}\vspace{-5mm}
\end{center}
\end{figure}

The goal for ASPiRe is to accelerate downstream task learning using a library of primitive skill priors. There are two main challenges for this task: 1) Decide which primitive skill prior should be activated given a new downstream task, where using unrelated primitive skill prior could mislead the learning process. 2) How to best combine multiple primitive skill priors, either sequentially or concurrently. In particular, to the best of our knowledge, how to simultaneously constrain the policy learning with multiple skill priors is an open challenge.

To address these challenges, we propose a method that regularizes the policy learning using an ``adaptive'' skill prior, which can be either one of the primitive skill priors or a newly created skill prior by the composition of multiple primitive skill priors (i.e., the composite skill prior).

Our approach consists of two learning stages: offline and online (See Fig. \ref{fig:overview}). In the offline stage, the system learns multiple primitive skill priors from a set of labeled datasets. Each dataset contains expert demonstrations of one specific primitive task. In the online stage (for downstream tasks), the algorithm creates an adaptive skill prior by assigning weight over primitive skill priors. The weights are generated by the adaptive weight module (AWM). Then they are used to regularize the policy training. In the running example in Fig. \ref{fig:overview}, the policy is regularized by avoid and navigation primitives \textbf{concurrently} when the obstacle is observed.

\textbf{Problem Formulation:}
In the offline learning stage, the algorithm has access to K datasets $\{\mathcal{D}_1,...,\mathcal{D}_K\}$ with task labelling $\{\mathcal{T}_0,..,\mathcal{T}_K\}$. Each dataset $\mathcal{D}_i$ contains action-state trajectories $\tau=\{(s_0,a_0),...,(s_T,a_T)\}$ for task $\mathcal{T}_i$ (i.e. push). The datasets are generated by expert behavior policy $\mu_1,..,\mu_K$. We further assume that none of the single behaviors can individually solve the typically more complex downstream learning tasks. 

The online learning stage for the downstream task is formulated as a reinforcement learning problem, modelled as an infinite-horizon Markov Decision Process (MDP) $\mathcal{M}=(\mathcal{S},\mathcal{A},\mathcal{P},\mathcal{R},\rho,\gamma)$, where $\mathcal{S}$ is the continuous state space, $\mathcal{A}$ is the continuous action space, $\mathcal{P}$ is the dynamic $p(s_{t+1}|s_t,a_t)$, $\mathcal{R}$ is the scalar reward $r_t$, $\rho$ is the initial state distribution and $\gamma$ is the discount factor. The goal is to learn a policy $\pi(a_t|s_t)$ to maximize the expected discounted sum of rewards $\sum_{t=0}^T \mathbb{E}_\pi[\gamma^t r(s_t,a_t)$].

\subsection{Extracting multiple primitive skill Priors} \label{sec:approach-skill}
We denote an action sequence $\{a_t,...,a_{t+H-1}\}$ with fixed horizon $H$ as a skill $\actionseq_t$. In the offline learning phase, we first learn a shared low-dimensional skill embedding space $\mathcal{Z}$ from aggregated datasets ${\{\mathcal{D}_i\}_{i=0}^K}$. We then construct primitive skill priors on top of the skill embedding space for each dataset $\mathcal{D}_i$. Distinct primitive skill priors can generate different types of skills based on the task associated with $\mathcal{D}_i$.   

To learn the skill embedding, following the previous works \cite{ajay2020opal,pertsch2020spirl,wang2022skill}, we fit a variational autoencoder and  maximize the following evidence lower bound (ELBO):
\begin{align}
    &\log p(\actionseq_t) \geq \mathbb{E}_q \bigg[\underbrace{\log p(\actionseq_t \vert z)}_{\text{reconstruction}} - \beta \big(\underbrace{\log q(z \vert \actionseq_t) - \log p(z)}_{\text{regularization}}\big) \bigg].
\end{align}
Here, $\beta$ is a parameter that controls the regularization term. The skill encoder $q(z \vert \actionseq_t)$ embeds the skill $\actionseq_t$ into latent space $\mathcal{Z}$ and  the skill decoder $p(\actionseq_t \vert z)$ reconstructs the latent vectors back to skills. We set $p(z)$ as a unit Gaussian prior to regularize the encoder learning. Note that action sequences $\actionseq$ are randomly sampled from all offline dataset ${\{\mathcal{D}_i\}_{i=0}^K}$ such that diverse action sequences can be embedded in a shared skill embedding space with the same encoder and decoder.  

We freeze the skill encoder and decoder parameters after sufficient training. Next, we extract multiple primitive skill priors on top of the shared skill embedding. For each offline dataset $\mathcal{D}_i$, we extract a primitive skill prior $P_a^i(z|s_t)$, which is parameterized as a Gaussian distribution. To train primitives, we sample state and skill tuples $(s_t,\actionseq_t)$ from $\mathcal{D}_i$ and minimize the KL divergence between the skill posterior and the predicted prior $\mathbb{E}_{(s_t, \actionseq_t) \sim \mathcal{D}_i}  D_{\text{KL}}\big(q(z \vert \actionseq_t), p_a^i(z \vert s_t)\big)$, similar to previous work \cite{pertsch2020spirl}.

\subsection{Regularized reinforcement learning via adaptive skill prior} \label{sec:approach-regularize}
For downstream task learning, instead of learning a policy on action space $\mathcal{A}$, we learn a policy $\pi_\theta(z|s_t)$ which acts on high-level skill embedding space $\mathcal{Z}$. Previous work \cite{pertsch2020spirl} has demonstrated that learning on the skill embedding space is beneficial for complex long-horizon tasks. The skill embedding $z$ can be decoded into an action sequence $\actionseq$ via the skill decoder $p(\actionseq \vert z)$ for $H$ steps action execution. We further replace the dynamic in our problem formulation with H-step dynamic $p'(s_{t+H}|s_t,z_t)$ and rewards with H-step rewards $r_t' = \sum_{t=1}^{H} r_t$. 

We guide the policy learning via an adaptive skill prior over the course of online learning for downstream tasks. Ideally, the policy should be regularized by all relevant primitive skill priors at each state. We consider two scenarios: 1) only one primitive prior is relevant, 2) multiple primitive priors are relevant. Previous work \cite{pertsch2020spirl} considers the first scenario and regularizes the learning policy by providing an additional negated KL divergence between the policy and the offline skill prior. Similarly, SVPG \citep{liu2017stein} regularizes the policy parameters by a prior distribution and optimize the object via SVGD \citep{liu2016stein}.

In states where multiple primitive skills are involved, the agent needs to achieve several goals simultaneously, which are implicitly encoded in distinct primitive skill priors. It is therefore natural to guide the policy learning with multiple primitive skill priors. To implement this idea, we can augment the standard RL objective with a negated \textit{weighted KL divergence}:

\begin{align}
    J(\pi)&= \sum_{t=0}^T \mathbb{E}_\pi\bigg[r(s_t,z_t)-\alpha \underbrace{\sum_{i=1}^K \omega_i(s_t)D_{KL}(\pi(z_t|s_t),p_a^i(z_t|s_t))\bigg]}_{\text{weighted KL divergence}}
\label{eq:augment object w multiple priors}
\end{align}

Here, we refer to $\omega(s)$ as the \textit{weighting function}, which specifies the weights $w = (w_1,w_2,...,w_K)$ to apply on KL divergence between the policy and primitive skill priors. We consider weights are bounded $\{w_i\}_{i=1}^{K}\in [0,1]$ and are normalized $\sum_{i=1}^{K} w_i=1$. Assigning weights to the primitive skill priors implicitly forms a \textit{composite skill prior}. This can be interpreted as the learned policy will be constrained by a skill distribution $p_c(z_t)$ that can minimize the weighted KL divergence given weights $w$ at state $s_t$:
\begin{align}
    p_c(z_t) = \argmin_{p_c}\sum_{i=1}^K w_i D_{KL}(p_c(z_t),p_a^i(z_t|s_t))
\end{align}
We refer to it as \textit{composite skill prior} generated by weight $w$ at state $s_t$. Weights at a given state can be viewed as a measure of how much one primitive skill prior should contribute to the composite skill prior. The state-dependent weights can be learned simultaneously with the policy, and we discuss it in section \ref{Sec: AWM}. For now, we assume the weighting function $\omega(s_t)$ is given, and the goal is to maximize the Equation \ref{eq:augment object w multiple priors}.

We follow the policy constraints formulation \cite{Kumar2019StabilizingOQ,DBLP:journals/corr/abs-1911-11361,siegel2019keep} to optimize the objective. We consider a Q-function $Q_\phi$ and a policy $\pi_\theta$ parameterized by the network with parameters $\phi$ and $\theta$, respectively. The policy parameters can be optimized by minimizing the weighted KL divergence penalty: 
\begin{align}
    J_{\pi}(\theta) &= \mathbb{E}_{s_t \sim D}\bigg[\mathbb{E}_{z_t \sim \pi_\theta}\big[-Q_\phi(s_t,z_t)+\alpha \sum_{i=1}^K \omega_i(s_t)D_{KL}(\pi_\theta(z_t|s_t),p_a^i(z_t|s_t))\big]\bigg]
    \label{eq:policy optim}
\end{align}
Here $\alpha\geq 0$ is a temperature parameter to constrain the weighted KL-divergence, which can be tuned automatically by providing a target divergence parameter $\delta$ \cite{pertsch2020spirl} (See appendix A.3 for details).

Q-function parameters can be optimized by minimizing the Bellman residual:
\begin{align}
    J_Q(\phi)&=\mathbb{E}_{s_t,z_t\sim D}\bigg[\frac{1}{2}\big(Q_\phi(s_t,z_t)-(r(s_t,z_t)+\gamma Q_{\Bar{\phi}}(s_{t+1},\pi_\theta(z_{t+1}|s_{t+1}))\big)^2\bigg]
\end{align}

\subsection{Adaptive Weight Module} \label{Sec: AWM}

In this section, we introduce Adaptive Weight Module (AWM), a component to infer the optimal weights adaptively over the course of tasks. Our goal is to learn a parameterized weighting function $\omega_\sigma$ to compose the primitive skill priors via weighted KL divergence, and the resulting composite skill prior can guide the policy learning. We evaluate the expected critic value of skills under the composite skill prior distribution generated by the weights $\omega$. The weights leading to the highest critic value is selected to assign among primitives. Both the weighting function $\omega_\sigma$ and the skill prior generator $G_\eta$ are learned online along with the policy $\pi_\theta$.

\textbf{Learning skill prior generator.} To efficiently evaluate the skills under composite skill priors, we implement skill prior generator $G_\eta(z|s_t,w)$, a generative model which outputs the composite skill prior generated by the weights $w$ given the state $s_t$. This can be optimized by minimizing the weighted KL divergence between output distribution $G_\eta(\cdot|s_t,w)$ and primitive skill priors:
\begin{equation}
    J_G(\eta) = \mathbb{E}_{s_t\sim\mathcal{D}}\sum_{i=1}^{K}w_i D_{KL}(G_\eta(z|s_t,w),P_a^i(z_{t}|s_{t}))
\end{equation}

We parameterize the skill prior generator as a Gaussian distribution by the network with parameters $\eta$. During the learning, we randomly sample weights $w$ from the unit simplex as the weight input. This helps the skill prior generator to learn all possible compositions across states.

\textbf{Learning adaptive skill prior.} Our goal is to construct an adaptive skill prior that is biased towards the high reward trajectories, similar to \citet{siegel2019keep}. The key difference is that our approach aims to construct such adaptive prior for a novel downstream task by composing multiple primitive skill priors. In effect, we would like to find weights assignment such that the corresponding composite skill prior can contain the skills that are more likely to lead to task success. We formulate this as:

\begin{align}
    \sigma = \argmax_\sigma \sum_t\mathbb{E}_{s_t \sim \mathcal{D}}\big[\mathbb{E}_{z \sim G_\eta(\cdot|s_t,\omega_\sigma(s_t))}[Q_\phi(s_t,z)]\big]
    \label{eq:awm_object}
\end{align}

where $Q_\phi$ is the policy critic defined in section \ref{sec:approach-regularize}. The Q-value in the objective measures the usefulness of skills for the task at hand. The inner expectation can be viewed as a measurement of the usefulness of the composite skill prior generated by weight $w$. The objective prefers a composite skill prior, in which skills have higher Q-value and are more promising to be explored. We approximate the inner expectation with $M$ samples from $G_\eta$, i.e, $\mathbb{E}_{z \sim G_\eta(\cdot|s_t,\omega_\sigma(s_t))}[Q_\phi(s_t,z)] = \frac{1}{M}[Q_\phi(s_t,z) \vert z\sim G_\eta(\cdot|s_t,\omega_\sigma(s_t))]$. We find $M=20$ is sufficient to provide a stable and accurate estimation. 

We parameterize the weighting function $\omega_\sigma$ by the network with parameter $\sigma$, and use softmax as the output layer. To learn the optimal weight, we set the Equation \ref{eq:awm_object} as the loss function. We add an additional regularization term to penalize weights for deviating from uniform weight $\frac{1}{K}$ in the early stage of the training. This encourages the learned policy to explore more skills in diverse primitive skill priors and prevents weights from collapsing to a singleton due to the noise in Q-value in the early phase. Here, $\beta$ is the annealing coefficient, and it gradually decreases. The weighting function parameters can be trained to minimize the following: 
\begin{align}
     J_{\omega}(\sigma) &= \mathbb{E}_{s_t \sim \mathcal{D}}\bigg[\mathbb{E}_{z \sim G_\eta(\cdot|s_t,\omega_\sigma(s_t))}\Big[-Q_{\phi}(s_t,z)+\beta||\omega_\sigma(s_t)-\frac{1}{K}||_2^2  \Big]\bigg]
    \label{eq:weight policy optim}
\end{align}

For each gradient step, we optimize both the policy and the AWM. The full Adaptive Skill Prior algorithm is summarized in Algorithm \ref{Algo:DP}. For implementation details, see appendix, A.4.

\begin{algorithm}[t]
\caption{ASPiRe Algorithm}
\label{Algo:DP}
\begin{algorithmic}[1]
\State \textbf{Inputs:} K primitive skill priors $\{P_a^i(z_t|s_t\}_{i=0}^{K}$
\State \textbf{Initialize:} replay buffer $\mathcal{D}$, policy $\pi_\theta(z_t|s_t)$, critic $Q_{\phi}(s_t,z_t)$, target critic  $Q_{\Bar{\phi}}(s_t,z_t)$, weighting function $\omega_\sigma(s_t)$, skill prior generator $G_\eta(z_t|s_t,w_t)$ 
\For{each iteration}
    \For{every H environment steps}
        \State $z_t \sim \pi_\theta(z_t|s_t)$  \Comment{Sample skill from policy}
        \State $s_t' \sim p(s_{t+H}|s_t,z_t)$. \Comment{Execute skill}
        \State $\mathcal{D}\leftarrow \mathcal{D} \cup \{s_t,z_t,r_t,s_t'\}$ \Comment{Store experience in replay buffer}
    \EndFor
    \For{each gradient step}
        \State Sample a batch of transition tuples $\{s_t,z_t,r_t,s_t'\}$ from $\mathcal{D}$ 
        \State $w_t = \omega_\sigma(s_t)$   \Comment{Get weights from weighting function}
        \State $\theta \leftarrow \theta -\lambda_\theta\nabla _\theta J_\pi(\theta)$ \Comment{Update policy parameters via weighted KL divergence}
        \State $\phi \leftarrow \phi -\lambda_\phi\nabla _\phi J_Q(\phi)$ \Comment{Update critic parameters}
        \State $\eta \leftarrow \eta -\lambda_\eta\nabla _\sigma J(\eta)$ \Comment{Update generator parameters}
        \State $\sigma \leftarrow \sigma -\lambda_\sigma\nabla _\sigma J(\sigma)$ \Comment{Update weighting parameters}
        \State $\Bar{\phi}\leftarrow \tau\phi +(1-\tau)\bar{\phi}$ \Comment{Update target critic}
    \EndFor
\EndFor
\end{algorithmic}
\end{algorithm}

\section{Experiments} \label{sec:experiment}
We would like to answer the following questions through our experiments: 
a) Can leveraging multiple primitive skill priors accelerate downstream task learning? 
b) What's the advantage of using the composite skills as regularization compared to directly using them as policy?
c) Can AWM optimally assign proper weights over different primitive skill priors based on the downstream tasks, even when irrelevant primitive skill priors are presented?

\subsection{Experiment setup}

We evaluate our method in three modified environments from D4RL \citep{fu2020d4rl}. ASPiRe is given two primitive skill priors in the first two experiments. For the last experiment, we test ASPiRe's ability to compose three primitive skill priors. For each environment, we first collect datasets for distinct skills from expert demonstrations, which allows us to extract distinct primitive skill priors. We then evaluate our method with extracted primitives in downstream tasks.

\textbf{Point Maze.} We modified the Point Maze environment by adding multiple obstacles in the maze. The task is to navigate a point mass agent in the maze and reach the goal. During this process, the agent must avoid the obstacles randomly placed in the maze. The original point maze environment is already challenging because of the sparse reward feedback. Randomly placing obstacles in the maze makes this problem even more difficult. For each episode, the maze layout is fixed. However, the agent start, goal, and obstacles locations are randomly specified. Our method will be equipped with two primitive skill priors for this task: navigation and avoid.

\textbf{Ant Push.} The ant needs to push the box to the goal while avoiding an obstacle in this task. ASPiRe will carry two primitive skill priors: push and avoid. The ant, box, obstacle, and goal locations are randomly generated at the beginning of every episode. The agent will receive a sparse reward upon the box reaching the goal location.

\textbf{Ant Maze.} We modified the Ant Maze by adding an obstacle and a box in the maze. The agent's goal is to traverse the maze while pushing the box to the goal position without hitting the obstacle. The agent always starts in the lower right corner of the maze. However, the box, goal, and obstacle positions are randomly selected from a pre-defined set. Ant Maze itself poses a complex exploration problem due to sparse reward feedback. We make this problem even more complex by changing the success criteria. In this case, ASPiRe will carry three primitives: push, navigation, and avoid.

\textbf{Robotic Manipulation.} We further test ASPiRe's ability in complex robotic manipulation task in robosuite environment \citep{robosuite2020}. The task is to control a Panda robot arm to grasp the box (in red) without colliding with the barrier (in green) placed in the middle of the desk (See sub-figure 1 in Fig. \ref{fig:robot env demo}).The location of the box and barrier are randomly generated at the beginning of the episode. ASPiRe will carry two primitive skill priors: grasp and avoid. 

See appendix A.4 for details on environment, data collection process and training. 

\begin{figure}[h]
\begin{center}
\includegraphics[width=\columnwidth]{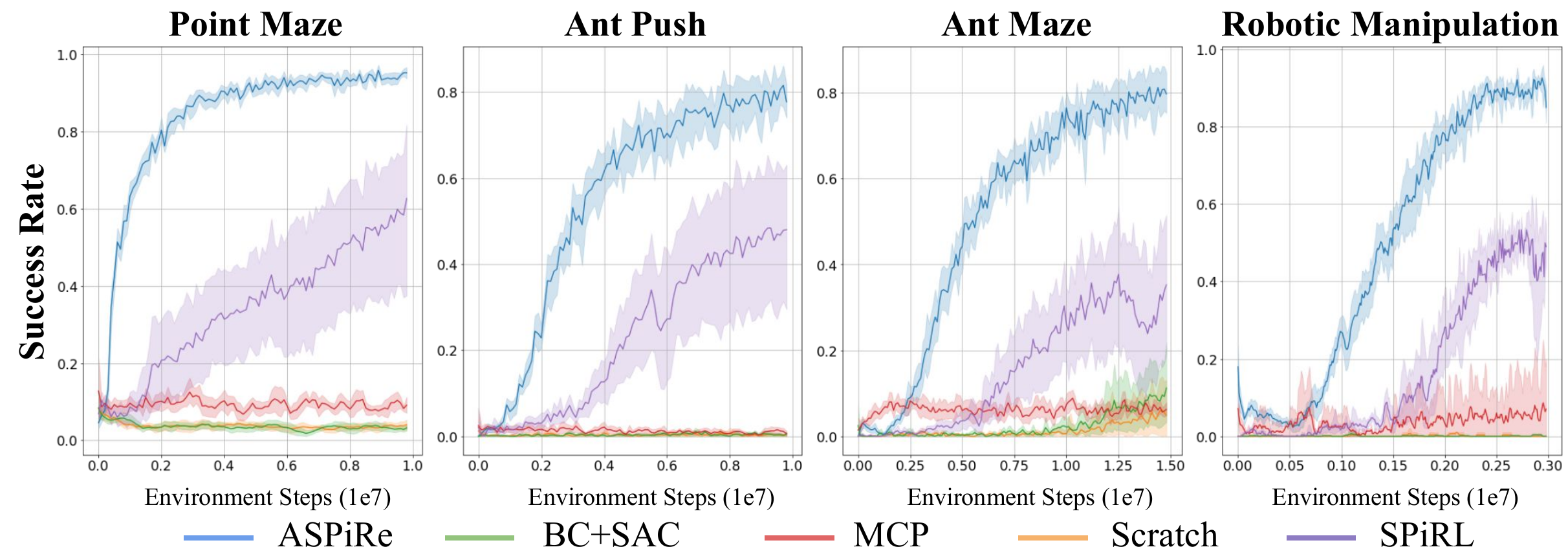}
\vspace{-3mm}
\caption{\textbf{Comparative Experiment.} Learning curves of our method and comparative methods for downstream tasks. ASPiRe significantly improves the learning efficiency and performance on all challenging tasks. The shaded area is the standard deviation over eight seeds.}
\label{fig:exp_comparsion}
\end{center}
\vspace{-2mm}
\end{figure}

\textbf{Alternative approaches.}
We compare ASPiRe with several prior works for downstream task learning performance. To fairly evaluate the performance, we let all methods (besides SPiRL, as the method learns its own skill embedding) act on skill embedding space $\mathcal{Z}$ learned by the ASPiRe.  
\begin{itemize}[leftmargin=6mm]
    \item \textbf{Scratch.} Train a Soft Actor-Critic (SAC, \citep{haarnoja2018soft}) agent from scratch. This comparison tests the benefit of using the learned skill priors to guide downstream task learning. 
    \item \textbf{Behavioral Cloning with finetuning.} A behavior cloning policy is first trained with offline data and then finetuned using SAC for the downstream tasks.  
    \item \textbf{MCP}\citep{Peng2019MCPLC}\textbf{.} The method composes primitive policies through multiplicative composition of Gaussians with a learnable weighting function. MCP then samples the actions from the resulting multiplicative composite distribution. For a fair evaluation, MCP is directly given the same set of primitives that ASPiRe is equipped with. This comparison tests the benefit of treating composite skills as prior, instead of directly executing them.  
    \item \textbf{SPiRL}
    \citep{pertsch2020spirl}\textbf{.} Guide the policy learning with a single skill prior. We train the skill prior by combining all primitive datasets into one and learning a single prior from it. This comparison tests the benefit of guiding the learning policy with multiple primitive skill priors.
    
\end{itemize}

\subsection{Experiment results}
As shown in Fig. \ref{fig:exp_comparsion}, our experiments demonstrate that ASPiRe is able to learn efficiently and converge to high success rates in all challenging environments.

\textbf{Benefits of multiple primitive skill priors.} 
We first compare ASPiRe with methods without skill prior guiding, namely training from scratch and behavior cloning with finetuning. They fail to solve all three tasks (Fig.  \ref{fig:exp_comparsion}). This confirms the finding in \citet{pertsch2020spirl} that skill prior is beneficial for learning complex and long-horizon tasks. Next, we compare our method with SPiRL \citep{pertsch2020spirl}, which regularizes the learned policy with a single skill prior extracted from the aggregated dataset. By leveraging the skill prior, it can achieve more efficient exploration but still relatively low learning speed and final success rates compared to ASPiRe (Fig.  \ref{fig:exp_comparsion}). We hypothesize that the performance gap between ASPiRe and SPiRL comes from the prior they deploy to guide the policy learning. We find that SPiRL learns an "average" behavior over all primitives. It prevents SPiRL from guiding states where the specific primitive skills are required. In the case of ASPiRe, it can construct an adaptive skill prior from all primitive skills priors. Therefore, the learned policy can receive sufficient guidance to explore and learn the tasks.    
 
\begin{figure}[h]
\begin{center}
\includegraphics[width=\columnwidth]{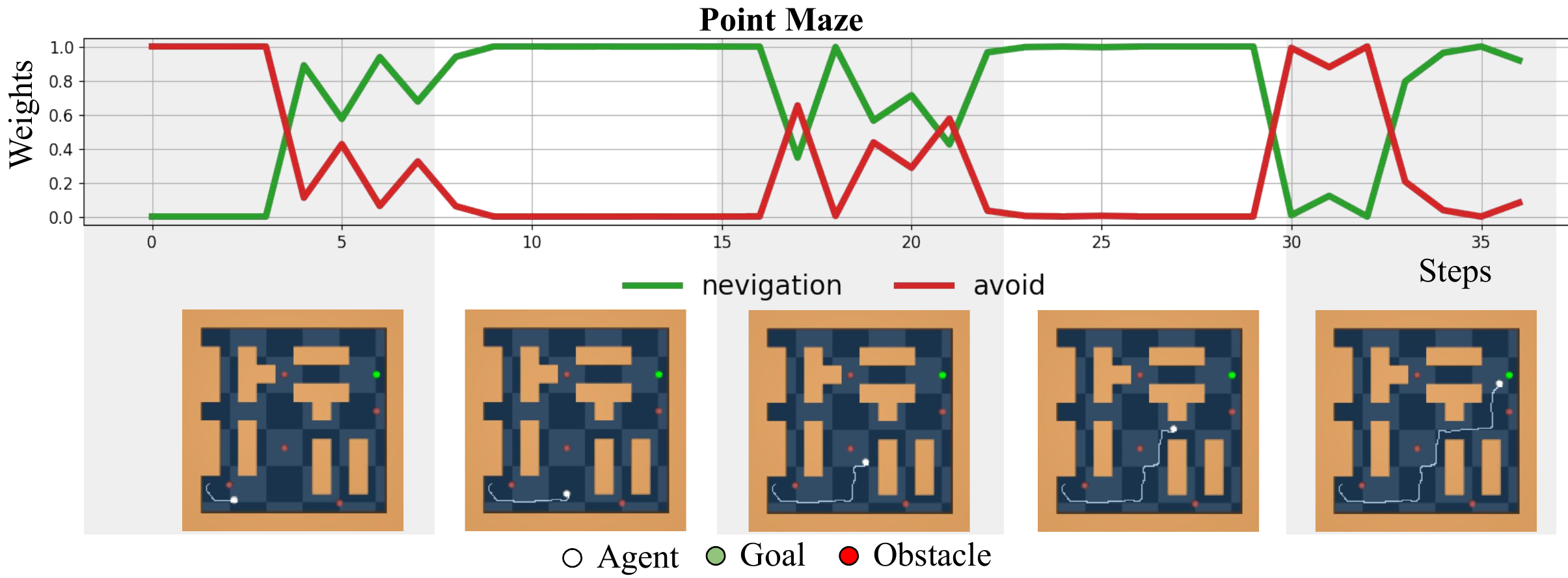}\vspace{-3mm}
\caption{\textbf{Weights inference over one episode (Point Maze)}. ASPiRe is able to compose the navigation prior and avoid prior concurrently (grey phases) when the obstacle is observed and only activate navigation prior when it is not (white phases).}
\label{fig:point maze demo}
\end{center}
\vspace{-3mm}
\end{figure}
\begin{figure}[h]
\begin{center}
\includegraphics[width=\columnwidth]{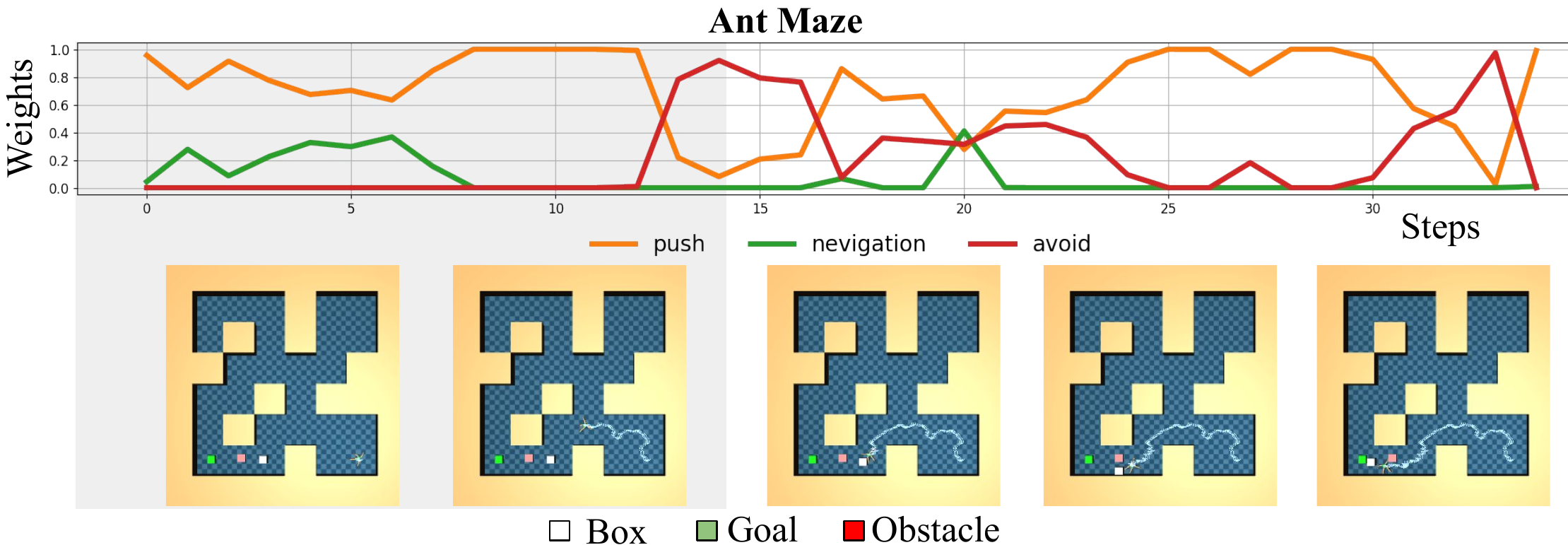} \vspace{-3mm}
\caption{\textbf{Weights inference over one episode (Ant Maze)} In the grey phase, ASPiRe is able to compose the navigation prior and push prior concurrently to traverse the maze and reach the box. In the white phase, the model composes the push and avoid prior to push the box to the goal while avoiding the obstacle along the way.}
\label{fig:ant maze demo}
\vspace{-5mm}
\end{center}
\end{figure}

\begin{figure}[h!]
\begin{center}
\includegraphics[width=\columnwidth]{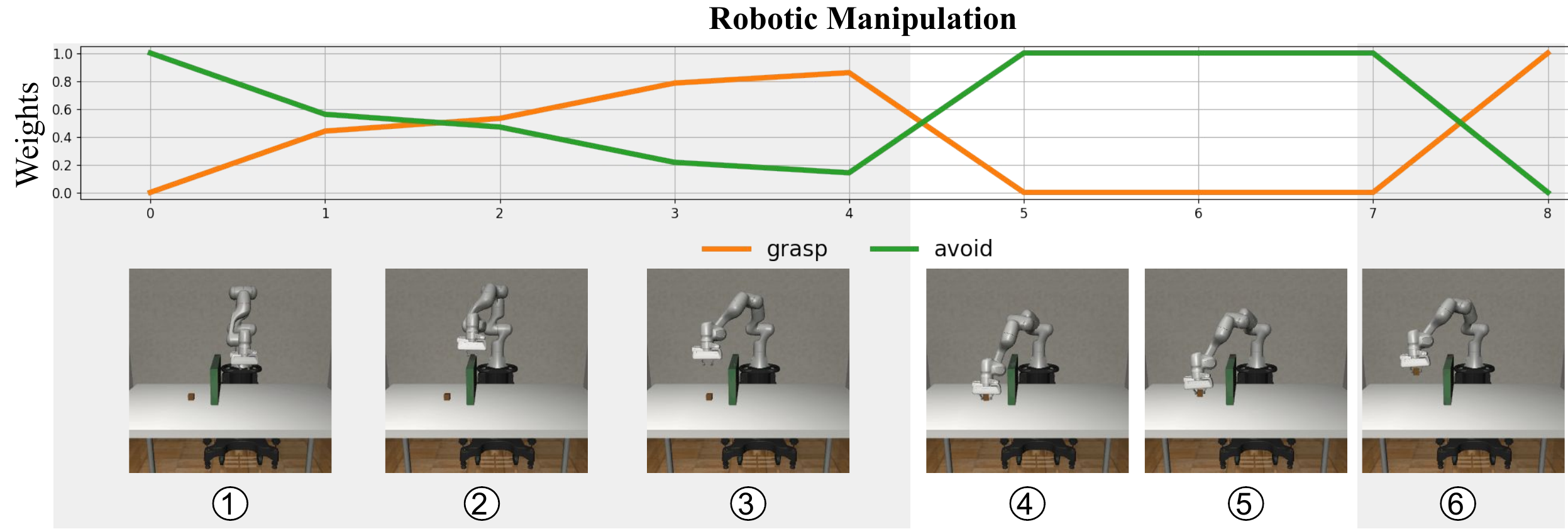} \vspace{-3mm}
\caption{\textbf{Weights inference over one episode (Robotic Manipulation)} In the first grey phase (sub-figure 1, 2, 3), ASPiRe is able to compose the grasp prior and avoid prior concurrently to lift the robot arm to avoid the barrier while approach to the box. In the white phase, notice that the robot arm is initially very close to the barrier (sub-figure 4). To avoid any potential collision between the robot arm and the barrier, ASPiRe activates the avoid prior to move the robot arm away from the barrier (sub-figure5). In the end, ASPiRe activates the grasp prior to grasp the box.}
\label{fig:robot env demo}
\vspace{-5mm}
\end{center}
\end{figure}

\textbf{Benefits of using composite skills as regularization.} 
We compare ASPiRe with MCP, which directly executes the composite skills as a policy with a learnable weight. 
As shown in Figure \ref{fig:exp_comparsion}, MCP struggles to learn a meaningful policy for all three tasks. 
This finding demonstrates that using composite skills as regularization provides the necessary flexibility for ASPiRe to learn the policies that can specifically solve the downstream task. We find the final learned policy has lower entropy than the skill priors in most cases. In fact, skill priors provide guidance on the valid skills to explore the environment, but the learned policy needs to further search in this valid skill space to solve the specific problem. We hypothesize that the resulting policy from MCP might be overly broad as the multiplicative composition of Gaussian inherent the relative high entropy from the primitives. 

\begin{wrapfigure}{r}{0.45\textwidth}\vspace{-3mm}
\begin{center}
\includegraphics[width=0.45\textwidth]{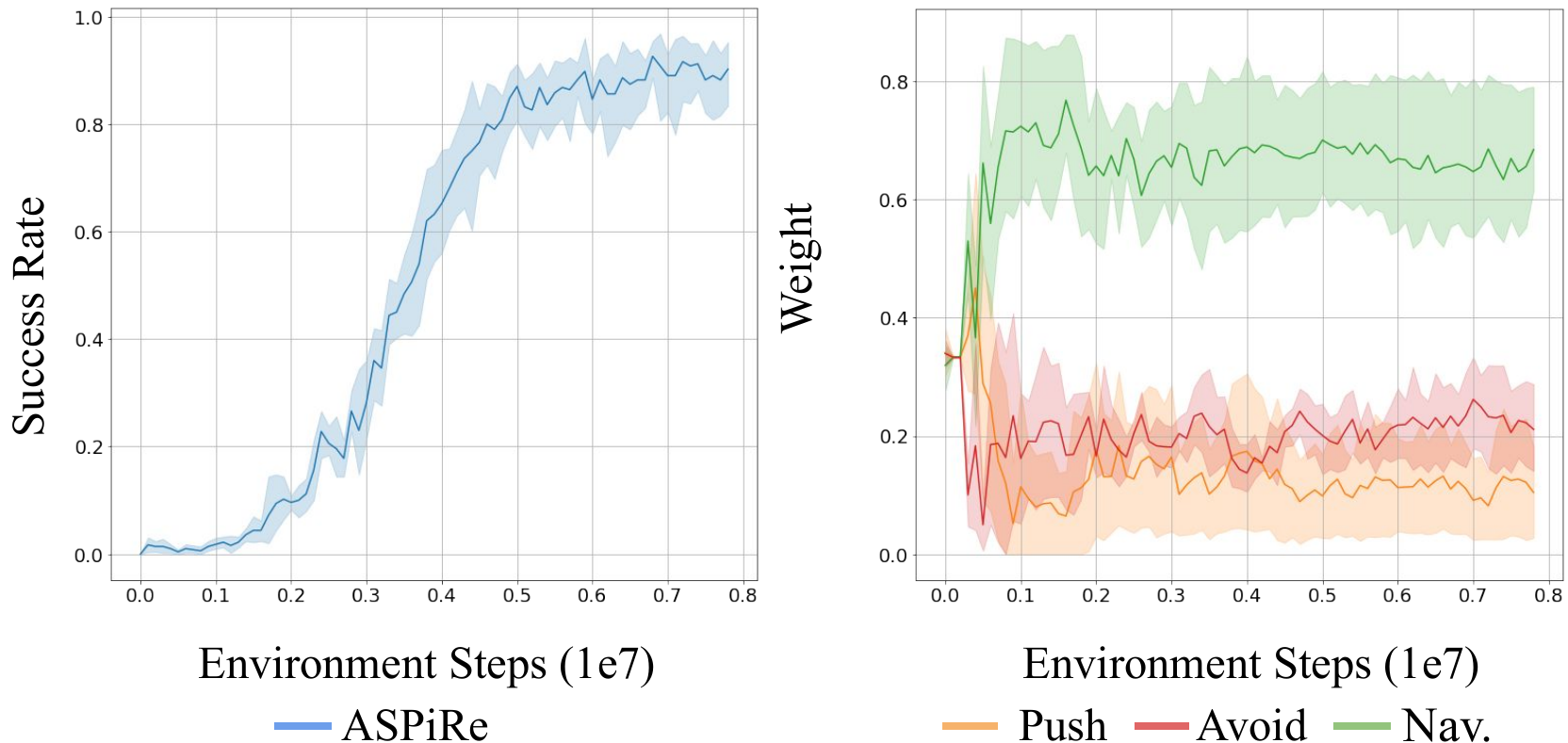}\vspace{-5mm}
\caption{\textbf{Training with unrelated primitive.} Left: Learning curve. Right: Average weights of skill priors, where we can observe that ASPiRe assigns a lower weight for unrelated skill prior (e.g., push).
}
\label{fig:ab2}
\end{center}\vspace{-4mm}
\end{wrapfigure}

\textbf{Learning optimal weights.} To demonstrate ASPiRe's ability to learn optimal weights, we visualize the weights generated by the weighting function for the primitive skills priors at different states for point maze, ant maze and robotic manipulation environments. From Fig. \ref{fig:point maze demo}, \ref{fig:ant maze demo} and \ref{fig:robot env demo}, we find that ASPiRe can compose the primitive skill priors both sequentially and concurrently throughout the tasks. More importantly, ASPiRe regularizes the learned policy with relevant priors, e.g., activates the avoid primitive when obstacles are near the agent. See appendix A.1 for more trajectories. 
\begin{wrapfigure}{r}{0.45\textwidth}\vspace{1mm}
\begin{center}
\includegraphics[width=0.45\textwidth]{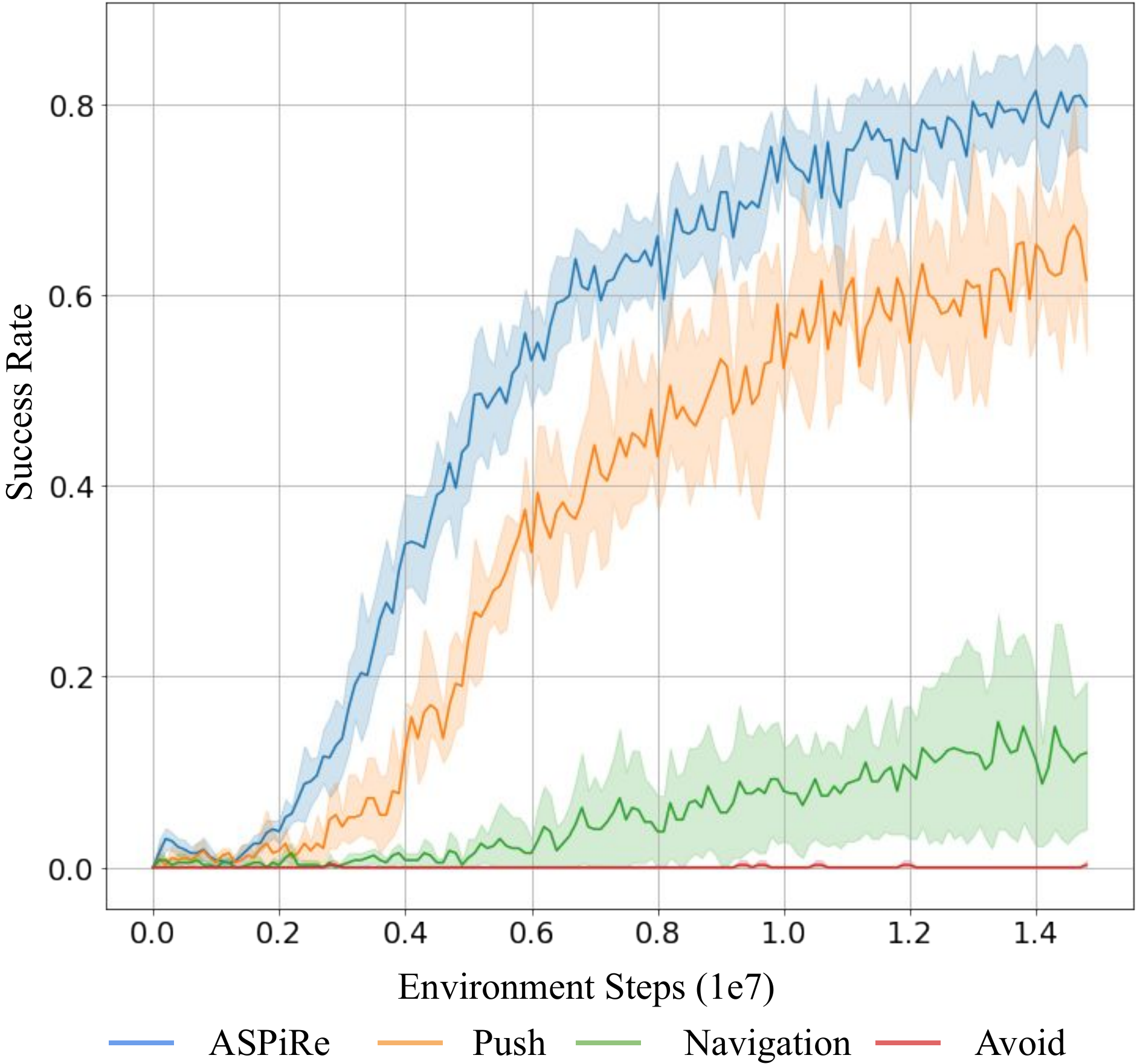}\vspace{-6mm}
\caption{\textbf{Guiding with composite or single primitive.} Learning curves in Ant Maze.}
\vspace{-5mm}
\label{fig:ablation}
\end{center}
\end{wrapfigure}

\textbf{Training with irrelevant primitive.}
To test ASPiRe's ability in handling irrelevant primitive skill prior, we modify the task of the Ant Maze environment, where the ant needs to reach the goal while avoiding the obstacle along the way without pushing the box. We place the box in the lower right corner of the maze, which is not on the path to the goal. In this case, the push primitive is irrelevant, and regularizing with it will yield low learning efficiency. From Fig. \ref{fig:ab2}, we find that ASPiRe still achieves good performance even with irrelevant primitive. We plot average weights over the states during the training. ASPiRe weights the navigation primitive heavily and assigns a low weight for the irrelevant primitive -- push. The weight for avoid is also relatively low since there are fewer states where avoiding is needed (i.e., only when the obstacle is in view). 



\textbf{Guiding with composite or single primitive.}
In this experiment, we test the advantage of using composite skill prior to guiding policy learning compared to guiding with a single primitive skill prior. 
To answer the question, we guide the policy learning with one of the primitive skill priors (i.e., push, navigation and avoid respectively) in ant maze environment. From Fig. \ref{fig:ablation}, we find that the learned policy guided with only the avoid primitive or navigation primitive fails to learn the task. Guided with the push primitive achieves relatively good performance but fails to complete the setting requiring traverse in the maze. This experiment suggests that composing multiple skill priors is beneficial for the complex multi-task downstream tasks. 

\textbf{Limitation and future work.}
When extracting primitive skill priors, our algorithm requires separated and labeled datasets, each containing a specialized skill. However, such dataset may not always be easy to generate, and manually separating or defining different primitive skill priors can be challenging and sometimes ambiguous. Therefore, an interesting future direction will be allowing the system to process unlabeled dataset and automatically extract a set of distinct primitive skill priors.

\section{Conclusion}
We present ASPiRe, an approach to transfer a spectrum of skills from offline data to accelerate the learning of unseen downstream tasks. We propose to guide the policy learning with multiple primitive skill priors and transfer the diverse skills to RL via weighed KL divergence. We model an Adaptive Weight Module to learn optimal weights assignment and construct an adaptive skill prior. We evaluate ASPiRe in challenging, sparse reward environments and demonstrate the benefits of guiding the learning with multiple skill priors. We also show that ASPiRe is able to select the most relevant skill priors even when unrelated skill priors are presented.

\section*{Acknowledgement}
The Authors would like to thank Nelson Vadori, Sumitra Ganesh, Huy Ha, Cheng Chi, Samir Gadre, Zhenjia Xu, and Zhanpeng He on
valuable feedback and support. 
Mengda Xu's work is supported by JPMorgan Chase \& Co. This paper was prepared for information purposes in part by the Artificial Intelligence Research group of JPMorgan Chase \& Co and its affiliates (“JP Morgan”), and is not a product of the Research Department of JP Morgan. JP Morgan makes no representation and warranty whatsoever and disclaims all liability, for the completeness, accuracy or reliability of the information contained herein. This document is not intended as investment research or investment advice, or a recommendation, offer or solicitation for the purchase or sale of any security, financial instrument, financial product or service, or to be used in any way for evaluating the merits of participating in any transaction, and shall not constitute a solicitation under any jurisdiction or to any person, if such solicitation under such jurisdiction or to such person would be unlawful.

\bibliographystyle{unsrtnat}
\bibliography{references}

\begin{thebibliography}{51}
\providecommand{\natexlab}[1]{#1}
\providecommand{\url}[1]{\texttt{#1}}
\expandafter\ifx\csname urlstyle\endcsname\relax
  \providecommand{\doi}[1]{doi: #1}\else
  \providecommand{\doi}{doi: \begingroup \urlstyle{rm}\Url}\fi

\bibitem[Ajay et~al.(2020)Ajay, Kumar, Agrawal, Levine, and
  Nachum]{ajay2020opal}
Anurag Ajay, Aviral Kumar, Pulkit Agrawal, Sergey Levine, and Ofir Nachum.
\newblock Opal: Offline primitive discovery for accelerating offline
  reinforcement learning.
\newblock In \emph{International Conference on Learning Representations}, 2020.

\bibitem[Pertsch et~al.(2020)Pertsch, Lee, and Lim]{pertsch2020spirl}
Karl Pertsch, Youngwoon Lee, and Joseph~J. Lim.
\newblock Accelerating reinforcement learning with learned skill priors.
\newblock In \emph{Conference on Robot Learning (CoRL)}, 2020.

\bibitem[Singh et~al.(2020)Singh, Liu, Zhou, Yu, Rhinehart, and
  Levine]{singh2020parrot}
Avi Singh, Huihan Liu, Gaoyue Zhou, Albert Yu, Nicholas Rhinehart, and Sergey
  Levine.
\newblock Parrot: Data-driven behavioral priors for reinforcement learning.
\newblock In \emph{International Conference on Learning Representations}, 2020.

\bibitem[Levine et~al.(2020)Levine, Kumar, Tucker, and
  Fu]{DBLP:journals/corr/abs-2005-01643}
Sergey Levine, Aviral Kumar, George Tucker, and Justin Fu.
\newblock Offline reinforcement learning: Tutorial, review, and perspectives on
  open problems.
\newblock \emph{CoRR}, abs/2005.01643, 2020.
\newblock URL \url{https://arxiv.org/abs/2005.01643}.

\bibitem[Kumar et~al.(2019)Kumar, Fu, Tucker, and
  Levine]{Kumar2019StabilizingOQ}
Aviral Kumar, Justin Fu, G.~Tucker, and Sergey Levine.
\newblock Stabilizing off-policy q-learning via bootstrapping error reduction.
\newblock In \emph{NeurIPS}, 2019.

\bibitem[Wu et~al.(2019)Wu, Tucker, and
  Nachum]{DBLP:journals/corr/abs-1911-11361}
Yifan Wu, George Tucker, and Ofir Nachum.
\newblock Behavior regularized offline reinforcement learning.
\newblock \emph{CoRR}, abs/1911.11361, 2019.
\newblock URL \url{http://arxiv.org/abs/1911.11361}.

\bibitem[Liu et~al.(2019)Liu, Swaminathan, Agarwal, and
  Brunskill]{DBLP:journals/corr/abs-1904-08473}
Yao Liu, Adith Swaminathan, Alekh Agarwal, and Emma Brunskill.
\newblock Off-policy policy gradient with state distribution correction.
\newblock \emph{CoRR}, abs/1904.08473, 2019.
\newblock URL \url{http://arxiv.org/abs/1904.08473}.

\bibitem[Fujimoto et~al.(2019)Fujimoto, Meger, and
  Precup]{Fujimoto2019OffPolicyDR}
Scott Fujimoto, David Meger, and Doina Precup.
\newblock Off-policy deep reinforcement learning without exploration.
\newblock In \emph{ICML}, 2019.

\bibitem[Kumar et~al.(2020)Kumar, Zhou, Tucker, and
  Levine]{kumar2020conservative}
Aviral Kumar, Aurick Zhou, George Tucker, and Sergey Levine.
\newblock Conservative q-learning for offline reinforcement learning.
\newblock \emph{Advances in Neural Information Processing Systems},
  33:\penalty0 1179--1191, 2020.

\bibitem[Siegel et~al.(2019)Siegel, Springenberg, Berkenkamp, Abdolmaleki,
  Neunert, Lampe, Hafner, Heess, and Riedmiller]{siegel2019keep}
Noah Siegel, Jost~Tobias Springenberg, Felix Berkenkamp, Abbas Abdolmaleki,
  Michael Neunert, Thomas Lampe, Roland Hafner, Nicolas Heess, and Martin
  Riedmiller.
\newblock Keep doing what worked: Behavior modelling priors for offline
  reinforcement learning.
\newblock In \emph{International Conference on Learning Representations}, 2019.

\bibitem[Wang et~al.(2020)Wang, Novikov, {\.Z}o{\l}na, Springenberg, Reed,
  Shahriari, Siegel, Merel, Gulcehre, Heess, et~al.]{wang2020critic}
Ziyu Wang, Alexander Novikov, Konrad {\.Z}o{\l}na, Jost~Tobias Springenberg,
  Scott Reed, Bobak Shahriari, Noah Siegel, Josh Merel, Caglar Gulcehre,
  Nicolas Heess, et~al.
\newblock Critic regularized regression.
\newblock In \emph{Proceedings of the 34th International Conference on Neural
  Information Processing Systems}, pages 7768--7778, 2020.

\bibitem[Agarwal et~al.(2019)Agarwal, Schuurmans, and
  Norouzi]{DBLP:journals/corr/abs-1907-04543}
Rishabh Agarwal, Dale Schuurmans, and Mohammad Norouzi.
\newblock Striving for simplicity in off-policy deep reinforcement learning.
\newblock \emph{CoRR}, abs/1907.04543, 2019.
\newblock URL \url{http://arxiv.org/abs/1907.04543}.

\bibitem[Jaques et~al.(2019)Jaques, Ghandeharioun, Shen, Ferguson, Lapedriza,
  Jones, Gu, and Picard]{DBLP:journals/corr/abs-1907-00456}
Natasha Jaques, Asma Ghandeharioun, Judy~Hanwen Shen, Craig Ferguson,
  {\`{A}}gata Lapedriza, Noah Jones, Shixiang Gu, and Rosalind~W. Picard.
\newblock Way off-policy batch deep reinforcement learning of implicit human
  preferences in dialog.
\newblock \emph{CoRR}, abs/1907.00456, 2019.
\newblock URL \url{http://arxiv.org/abs/1907.00456}.

\bibitem[Kidambi et~al.(2020)Kidambi, Rajeswaran, Netrapalli, and
  Joachims]{kidambi2020morel}
Rahul Kidambi, Aravind Rajeswaran, Praneeth Netrapalli, and Thorsten Joachims.
\newblock Morel: Model-based offline reinforcement learning.
\newblock \emph{Advances in neural information processing systems},
  33:\penalty0 21810--21823, 2020.

\bibitem[Yu et~al.(2020)Yu, Thomas, Yu, Ermon, Zou, Levine, Finn, and
  Ma]{yu2020mopo}
Tianhe Yu, Garrett Thomas, Lantao Yu, Stefano Ermon, James~Y Zou, Sergey
  Levine, Chelsea Finn, and Tengyu Ma.
\newblock Mopo: Model-based offline policy optimization.
\newblock \emph{Advances in Neural Information Processing Systems},
  33:\penalty0 14129--14142, 2020.

\bibitem[Argenson and Dulac-Arnold(2020)]{argenson2020model}
Arthur Argenson and Gabriel Dulac-Arnold.
\newblock Model-based offline planning.
\newblock In \emph{International Conference on Learning Representations}, 2020.

\bibitem[Matsushima et~al.(2020)Matsushima, Furuta, Matsuo, Nachum, and
  Gu]{matsushima2020deployment}
Tatsuya Matsushima, Hiroki Furuta, Yutaka Matsuo, Ofir Nachum, and Shixiang Gu.
\newblock Deployment-efficient reinforcement learning via model-based offline
  optimization.
\newblock In \emph{International Conference on Learning Representations}, 2020.

\bibitem[Cang et~al.(2021)Cang, Rajeswaran, Abbeel, and
  Laskin]{cang2021behavioral}
Catherine Cang, Aravind Rajeswaran, Pieter Abbeel, and Michael Laskin.
\newblock Behavioral priors and dynamics models: Improving performance and
  domain transfer in offline rl.
\newblock In \emph{Deep RL Workshop NeurIPS 2021}, 2021.

\bibitem[Nair et~al.(2020)Nair, Dalal, Gupta, and
  Levine]{DBLP:journals/corr/abs-2006-09359}
Ashvin Nair, Murtaza Dalal, Abhishek Gupta, and Sergey Levine.
\newblock Accelerating online reinforcement learning with offline datasets.
\newblock \emph{CoRR}, abs/2006.09359, 2020.
\newblock URL \url{https://arxiv.org/abs/2006.09359}.

\bibitem[Rao et~al.(2021)Rao, Sadeghi, Hasenclever, Wulfmeier, Zambelli,
  Vezzani, Tirumala, Aytar, Merel, Heess, et~al.]{rao2021learning}
Dushyant Rao, Fereshteh Sadeghi, Leonard Hasenclever, Markus Wulfmeier, Martina
  Zambelli, Giulia Vezzani, Dhruva Tirumala, Yusuf Aytar, Josh Merel, Nicolas
  Heess, et~al.
\newblock Learning transferable motor skills with hierarchical latent mixture
  policies.
\newblock In \emph{International Conference on Learning Representations}, 2021.

\bibitem[Fern\'{a}ndez and Veloso(2006)]{10.1145/1160633.1160762}
Fernando Fern\'{a}ndez and Manuela Veloso.
\newblock Probabilistic policy reuse in a reinforcement learning agent.
\newblock In \emph{Proceedings of the Fifth International Joint Conference on
  Autonomous Agents and Multiagent Systems}, AAMAS '06, page 720–727, New
  York, NY, USA, 2006. Association for Computing Machinery.
\newblock ISBN 1595933034.
\newblock \doi{10.1145/1160633.1160762}.
\newblock URL \url{https://doi.org/10.1145/1160633.1160762}.

\bibitem[Browning et~al.(2004)Browning, Bruce, Bowling, and Veloso]{article}
Brett Browning, James Bruce, Michael Bowling, and Manuela Veloso.
\newblock Stp: Skills, tactics, and plays for multi-robot control in
  adversarial environments.
\newblock \emph{Proceedings of The Institution of Mechanical Engineers Part
  I-journal of Systems and Control Engineering - PROC INST MECH ENG I-J SYST
  C}, 219, 12 2004.
\newblock \doi{10.1243/095965105X9470}.

\bibitem[Haarnoja et~al.(2018{\natexlab{a}})Haarnoja, Pong, Zhou, Dalal,
  Abbeel, and Levine]{DBLP:journals/corr/abs-1803-06773}
Tuomas Haarnoja, Vitchyr Pong, Aurick Zhou, Murtaza Dalal, Pieter Abbeel, and
  Sergey Levine.
\newblock Composable deep reinforcement learning for robotic manipulation.
\newblock \emph{CoRR}, abs/1803.06773, 2018{\natexlab{a}}.
\newblock URL \url{http://arxiv.org/abs/1803.06773}.

\bibitem[Hunt et~al.(2019)Hunt, Barreto, Lillicrap, and
  Heess]{Hunt2019ComposingEP}
Jonathan~J. Hunt, Andr{\'e} Barreto, Timothy~P. Lillicrap, and Nicolas
  Manfred~Otto Heess.
\newblock Composing entropic policies using divergence correction.
\newblock In \emph{ICML}, 2019.

\bibitem[Van~Niekerk et~al.(2019)Van~Niekerk, James, Earle, and
  Rosman]{pmlr-v97-van-niekerk19a}
Benjamin Van~Niekerk, Steven James, Adam Earle, and Benjamin Rosman.
\newblock Composing value functions in reinforcement learning.
\newblock In Kamalika Chaudhuri and Ruslan Salakhutdinov, editors,
  \emph{Proceedings of the 36th International Conference on Machine Learning},
  volume~97 of \emph{Proceedings of Machine Learning Research}, pages
  6401--6409. PMLR, 09--15 Jun 2019.
\newblock URL \url{https://proceedings.mlr.press/v97/van-niekerk19a.html}.

\bibitem[Barreto et~al.(2017)Barreto, Dabney, Munos, Hunt, Schaul, van Hasselt,
  and Silver]{barreto2017successor}
Andr{\'e} Barreto, Will Dabney, R{\'e}mi Munos, Jonathan~J Hunt, Tom Schaul,
  Hado~P van Hasselt, and David Silver.
\newblock Successor features for transfer in reinforcement learning.
\newblock \emph{Advances in neural information processing systems}, 30, 2017.

\bibitem[Barreto et~al.(2018)Barreto, Borsa, Quan, Schaul, Silver, Hessel,
  Mankowitz, Zidek, and Munos]{barreto2018transfer}
Andre Barreto, Diana Borsa, John Quan, Tom Schaul, David Silver, Matteo Hessel,
  Daniel Mankowitz, Augustin Zidek, and Remi Munos.
\newblock Transfer in deep reinforcement learning using successor features and
  generalised policy improvement.
\newblock In \emph{International Conference on Machine Learning}, pages
  501--510. PMLR, 2018.

\bibitem[Borsa et~al.(2018)Borsa, Barreto, Quan, Mankowitz, van Hasselt, Munos,
  Silver, and Schaul]{borsa2018universal}
Diana Borsa, Andre Barreto, John Quan, Daniel~J Mankowitz, Hado van Hasselt,
  Remi Munos, David Silver, and Tom Schaul.
\newblock Universal successor features approximators.
\newblock In \emph{International Conference on Learning Representations}, 2018.

\bibitem[Barreto et~al.(2019)Barreto, Borsa, Hou, Comanici, Ayg{\"u}n, Hamel,
  Toyama, Hunt, Mourad, Silver, and Precup]{Barreto2019TheOK}
Andr{\'e} Barreto, Diana Borsa, Shaobo Hou, Gheorghe Comanici, Eser Ayg{\"u}n,
  Philippe Hamel, Daniel Toyama, Jonathan~J. Hunt, Shibl Mourad, David Silver,
  and Doina Precup.
\newblock The option keyboard: Combining skills in reinforcement learning.
\newblock In \emph{NeurIPS}, 2019.

\bibitem[Bacon et~al.(2016)Bacon, Harb, and Precup]{bacon2016optioncritic}
Pierre-Luc Bacon, Jean Harb, and Doina Precup.
\newblock The option-critic architecture, 2016.

\bibitem[Hausknecht and Stone(2016)]{hausknecht2016deep}
Matthew Hausknecht and Peter Stone.
\newblock Deep reinforcement learning in parameterized action space, 2016.

\bibitem[Frans et~al.(2018)Frans, Ho, Chen, Abbeel, and
  Schulman]{frans2018meta}
Kevin Frans, Jonathan Ho, Xi~Chen, Pieter Abbeel, and John Schulman.
\newblock Meta learning shared hierarchies.
\newblock In \emph{International Conference on Learning Representations}, 2018.

\bibitem[Merel et~al.(2018{\natexlab{a}})Merel, Ahuja, Pham, Tunyasuvunakool,
  Liu, Tirumala, Heess, and Wayne]{merel2018hierarchical}
Josh Merel, Arun Ahuja, Vu~Pham, Saran Tunyasuvunakool, Siqi Liu, Dhruva
  Tirumala, Nicolas Heess, and Greg Wayne.
\newblock Hierarchical visuomotor control of humanoids.
\newblock In \emph{International Conference on Learning Representations},
  2018{\natexlab{a}}.

\bibitem[Wulfmeier et~al.(2019)Wulfmeier, Abdolmaleki, Hafner, Springenberg,
  Neunert, Hertweck, Lampe, Siegel, Heess, and
  Riedmiller]{DBLP:journals/corr/abs-1906-11228}
Markus Wulfmeier, Abbas Abdolmaleki, Roland Hafner, Jost~Tobias Springenberg,
  Michael Neunert, Tim Hertweck, Thomas Lampe, Noah~Y. Siegel, Nicolas Heess,
  and Martin~A. Riedmiller.
\newblock Regularized hierarchical policies for compositional transfer in
  robotics.
\newblock \emph{CoRR}, abs/1906.11228, 2019.
\newblock URL \url{http://arxiv.org/abs/1906.11228}.

\bibitem[Heess et~al.(2016)Heess, Wayne, Tassa, Lillicrap, Riedmiller, and
  Silver]{heess2016learning}
Nicolas Heess, Greg Wayne, Yuval Tassa, Timothy Lillicrap, Martin Riedmiller,
  and David Silver.
\newblock Learning and transfer of modulated locomotor controllers, 2016.

\bibitem[Florensa et~al.(2017)Florensa, Duan, and
  Abbeel]{DBLP:journals/corr/FlorensaDA17}
Carlos Florensa, Yan Duan, and Pieter Abbeel.
\newblock Stochastic neural networks for hierarchical reinforcement learning.
\newblock \emph{CoRR}, abs/1704.03012, 2017.
\newblock URL \url{http://arxiv.org/abs/1704.03012}.

\bibitem[Merel et~al.(2018{\natexlab{b}})Merel, Hasenclever, Galashov, Ahuja,
  Pham, Wayne, Teh, and Heess]{merel2018neural}
Josh Merel, Leonard Hasenclever, Alexandre Galashov, Arun Ahuja, Vu~Pham, Greg
  Wayne, Yee~Whye Teh, and Nicolas Heess.
\newblock Neural probabilistic motor primitives for humanoid control.
\newblock In \emph{International Conference on Learning Representations},
  2018{\natexlab{b}}.

\bibitem[Hausman et~al.(2018)Hausman, Springenberg, Wang, Heess, and
  Riedmiller]{hausman2018learning}
Karol Hausman, Jost~Tobias Springenberg, Ziyu Wang, Nicolas Heess, and Martin
  Riedmiller.
\newblock Learning an embedding space for transferable robot skills.
\newblock In \emph{International Conference on Learning Representations}, 2018.

\bibitem[Eysenbach et~al.(2018)Eysenbach, Gupta, Ibarz, and
  Levine]{eysenbach2018diversity}
Benjamin Eysenbach, Abhishek Gupta, Julian Ibarz, and Sergey Levine.
\newblock Diversity is all you need: Learning skills without a reward function.
\newblock In \emph{International Conference on Learning Representations}, 2018.

\bibitem[Haarnoja et~al.(2018{\natexlab{b}})Haarnoja, Hartikainen, Abbeel, and
  Levine]{Haarnoja2018LatentSP}
Tuomas Haarnoja, Kristian Hartikainen, P.~Abbeel, and Sergey Levine.
\newblock Latent space policies for hierarchical reinforcement learning.
\newblock In \emph{ICML}, 2018{\natexlab{b}}.

\bibitem[Jacobs et~al.(1991)Jacobs, Jordan, Nowlan, and Hinton]{6797059}
Robert~A. Jacobs, Michael~I. Jordan, Steven~J. Nowlan, and Geoffrey~E. Hinton.
\newblock Adaptive mixtures of local experts.
\newblock \emph{Neural Computation}, 3\penalty0 (1):\penalty0 79--87, 1991.
\newblock \doi{10.1162/neco.1991.3.1.79}.

\bibitem[Qureshi et~al.(2020)Qureshi, Johnson, Qin, Henderson, Boots, and
  Yip]{Qureshi2020ComposingTP}
A.~H. Qureshi, Jacob~J. Johnson, Yuzhe Qin, Taylor Henderson, Byron Boots, and
  Michael~C. Yip.
\newblock Composing task-agnostic policies with deep reinforcement learning.
\newblock In \emph{ICLR}, 2020.

\bibitem[Tseng et~al.(2021)Tseng, Lin, Feng, and Sun]{tseng2021toward}
Wei-Cheng Tseng, Jin-Siang Lin, Yao-Min Feng, and Min Sun.
\newblock Toward robust long range policy transfer.
\newblock In \emph{Proceedings of the AAAI Conference on Artificial
  Intelligence}, volume~35, pages 9958--9966, 2021.

\bibitem[Ren et~al.(2021)Ren, Li, Ding, Pan, and Dong]{ren2021probabilistic}
Jie Ren, Yewen Li, Zihan Ding, Wei Pan, and Hao Dong.
\newblock Probabilistic mixture-of-experts for efficient deep reinforcement
  learning, 2021.

\bibitem[Wang et~al.(2022)Wang, Lee, Hakhamaneshi, Abbeel, and
  Laskin]{wang2022skill}
Xiaofei Wang, Kimin Lee, Kourosh Hakhamaneshi, Pieter Abbeel, and Michael
  Laskin.
\newblock Skill preferences: Learning to extract and execute robotic skills
  from human feedback.
\newblock In \emph{Conference on Robot Learning}, pages 1259--1268. PMLR, 2022.

\bibitem[Liu et~al.(2017)Liu, Ramachandran, Liu, and Peng]{liu2017stein}
Yang Liu, Prajit Ramachandran, Qiang Liu, and Jian Peng.
\newblock Stein variational policy gradient.
\newblock In \emph{33rd Conference on Uncertainty in Artificial Intelligence,
  UAI 2017}, 2017.

\bibitem[Liu and Wang(2016)]{liu2016stein}
Qiang Liu and Dilin Wang.
\newblock Stein variational gradient descent: A general purpose bayesian
  inference algorithm.
\newblock \emph{Advances in neural information processing systems}, 29, 2016.

\bibitem[Fu et~al.(2020)Fu, Kumar, Nachum, Tucker, and Levine]{fu2020d4rl}
Justin Fu, Aviral Kumar, Ofir Nachum, George Tucker, and Sergey Levine.
\newblock D4rl: Datasets for deep data-driven reinforcement learning, 2020.

\bibitem[Zhu et~al.(2020)Zhu, Wong, Mandlekar, and
  Mart\'{i}n-Mart\'{i}n]{robosuite2020}
Yuke Zhu, Josiah Wong, Ajay Mandlekar, and Roberto Mart\'{i}n-Mart\'{i}n.
\newblock robosuite: A modular simulation framework and benchmark for robot
  learning.
\newblock In \emph{arXiv preprint arXiv:2009.12293}, 2020.

\bibitem[Haarnoja et~al.(2018{\natexlab{c}})Haarnoja, Zhou, Abbeel, and
  Levine]{haarnoja2018soft}
Tuomas Haarnoja, Aurick Zhou, Pieter Abbeel, and Sergey Levine.
\newblock Soft actor-critic: Off-policy maximum entropy deep reinforcement
  learning with a stochastic actor.
\newblock In \emph{International conference on machine learning}, pages
  1861--1870. PMLR, 2018{\natexlab{c}}.

\bibitem[Peng et~al.(2019)Peng, Chang, Zhang, Abbeel, and
  Levine]{Peng2019MCPLC}
Xue~Bin Peng, Michael Chang, Grace~H. Zhang, P.~Abbeel, and Sergey Levine.
\newblock Mcp: Learning composable hierarchical control with multiplicative
  compositional policies.
\newblock In \emph{NeurIPS}, 2019.

\end{thebibliography}
\appendix

\section{Appendix}

\subsection{Hyperparameter Sensitivity Analysis}
\label{sec:hyper}
\begin{figure}[h]
\includegraphics[width=\columnwidth]{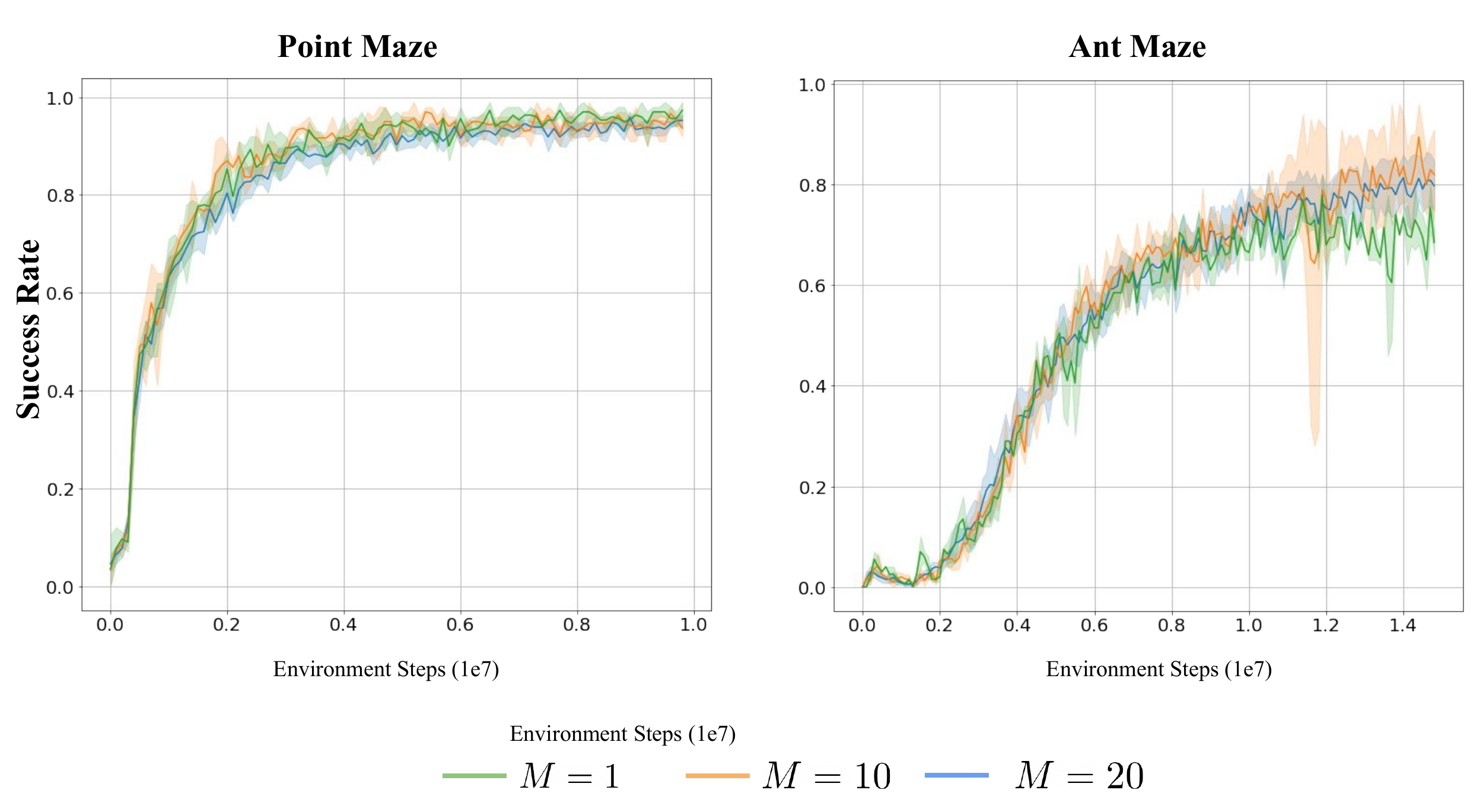}
\vspace{-3mm}
\caption{\textbf{Sensitivity towards sample size.} Learning curves of our method with different sample size $M$ in Point Maze and Ant Maze. The algorithm is not sensitive to this parameter. }
\label{fig:Hyper-M}
\includegraphics[width=\columnwidth]{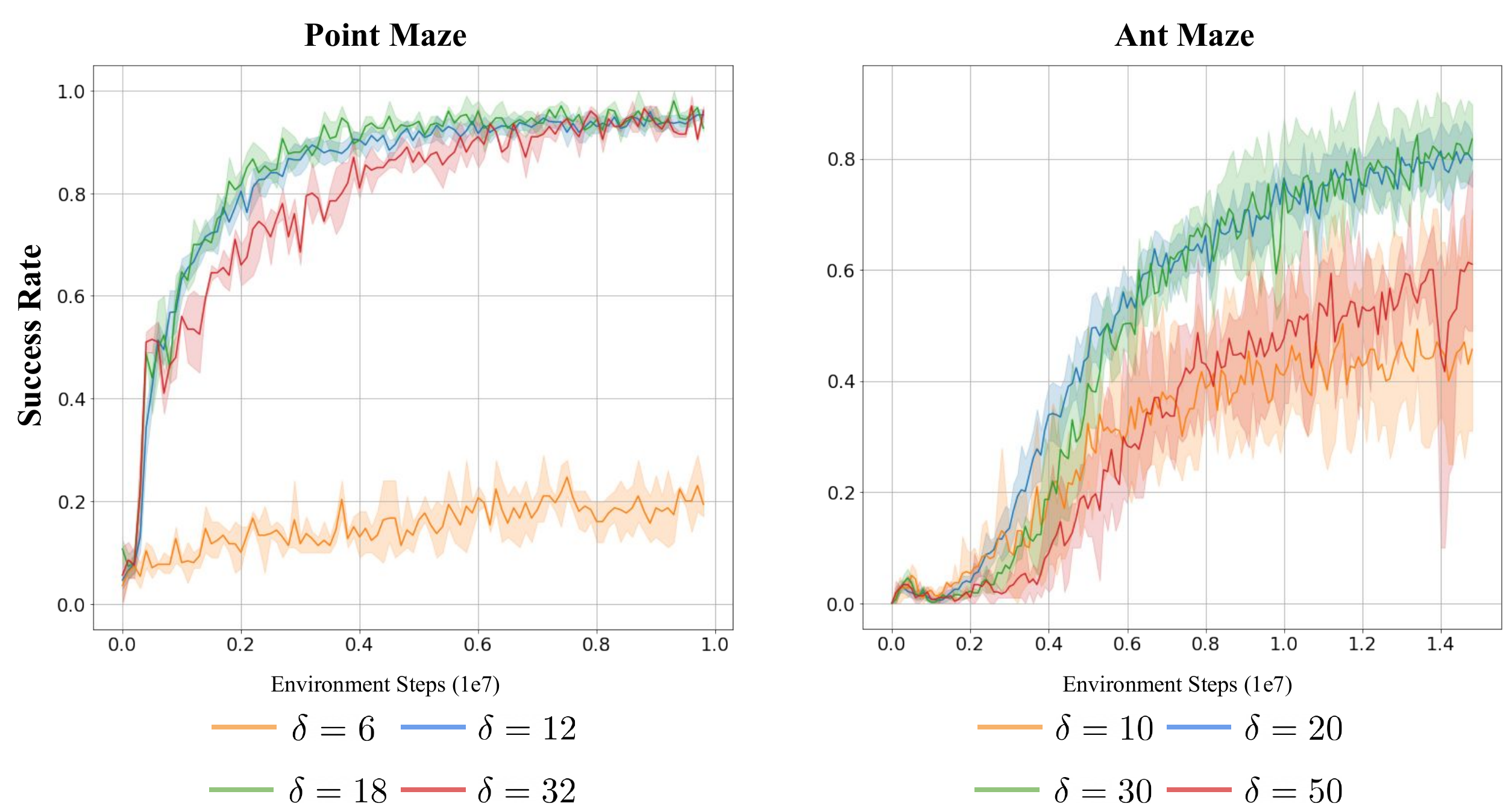}
\vspace{-3mm}
\caption{\textbf{Sensitivity towards target KL divergence.} Learning curves of our method with different target KL divergence $\delta$ in Point Maze and Ant Maze. More discussion in Sec \ref{sec:hyper}. }
\label{fig:hyper-kl}

\end{figure}

There are two hyperparameters that can potentially impact the online learning for the downstream task: a) The target KL divergence $\delta$, which is used to automatically tune the temperature parameters in Eq. \ref{eq:policy optim} and b) The sample size $M$ which is used to estimate the critic value in Eq. \ref{eq:awm_object}. We investigate the impact of the sample size $M$ on both Point Maze and Ant Maze environment and the choice of the target KL divergence $\delta$.

We first investigate the impact of sample size M by setting $M=1,10,20$. As shown in Fig. \ref{fig:Hyper-M}, We find that the sample size has almost no impact on the learning. We hypothesize two reasons: a) The large train batch size in our experiment helps to estimate the critic value $Q(s,\omega)$ more stable and b) Given the input size of weight vector $\omega$ is small, it might be easy for the neural network to approximate its value via small sample size.

We further investigate the impact of the target KL divergence $\delta$. As shown in Fig. \ref{fig:hyper-kl}, we find that target KL divergence significantly impacts the learning. Based on our experiments, we conclude that a) More complex tasks require higher target KL divergence. Notice that the target KL divergence imposes on Ant Maze is higher than the one on Point Maze. The optimal policy to solve complex tasks might be significantly different from the composite skill prior. Therefore, the policy needs more “space” to explore around the composite skill prior. b) Imposing too small target KL divergence can lead to downgraded performance. From Fig. \ref{fig:hyper-kl}, the ASPiRe's performance drops significantly when we set $\delta=6$ in Point Maze and $\delta=10$ in Ant Maze. This is because the policy will be forced to stay close to the composite skill prior and itself might not be able to solve such complex tasks. c) Imposing too big target KL divergence can lead to downgraded performance. We observe that the learning is not efficient when setting $\delta=32$ in Point Maze and $\delta=50$ in Ant maze. As target KL divergence increases, the learned policy will receive less guidance from the prior. Though the learning is sensitive to the choice of target KL divergence, we find that there might still be a range of KL divergences values leading to the same optimal performance. Notice the ASPiRe achieves the same good performance when setting $\delta=12$ or $\delta=18$ in the Point Maze. We can also observe the same behavior on Ant Maze.

\subsection{Composite Skill Prior as Policy}
\begin{figure}[h]
\begin{center}
\includegraphics[width=0.9\columnwidth]{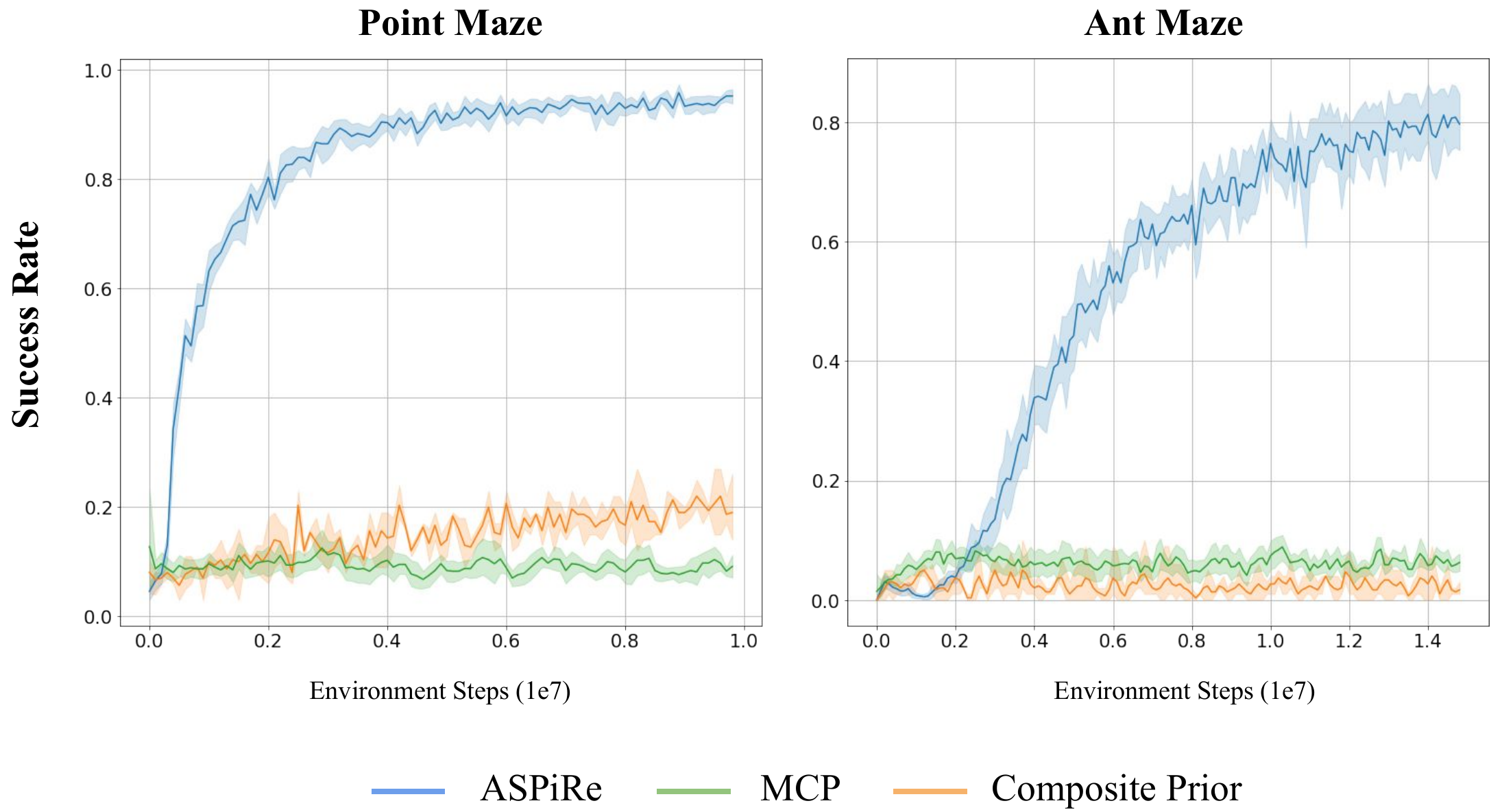}
\vspace{-3mm}
\caption{\textbf{Composite Skill Prior as Policy.} Learning curves of ASPiRe, MCP and Composite Skill Prior as policy in Point Maze and Ant Maze. Composite skill prior achieves similar performance as MCP which is much worse than ASPiRe.}
\label{fig:base-composite-skill-prior-policy}
\end{center}
\vspace{-6mm}
\end{figure}
We investigate the importance of using composite skill prior as regularization by comparing the ASPiRe with directly executing composite skill prior as a policy. This additional baseline is similar to the MCP. Both of them try to directly sample the action from composite distribution, but the difference is how they composite the distribution. One uses multiplicative gaussian (MCP) the other uses weighted KL divergence(ours).  

From Fig. \ref{fig:base-composite-skill-prior-policy}, we observe that using composite skill prior as a policy result in poor performance (similar performance as MCP). This confirms the importance of using composite skill prior as prior instead of policy. 

\subsection{Uniform Weights}
\label{sec:uniform weights}
\begin{figure}[h]
\begin{center}
\includegraphics[width=0.9\columnwidth]{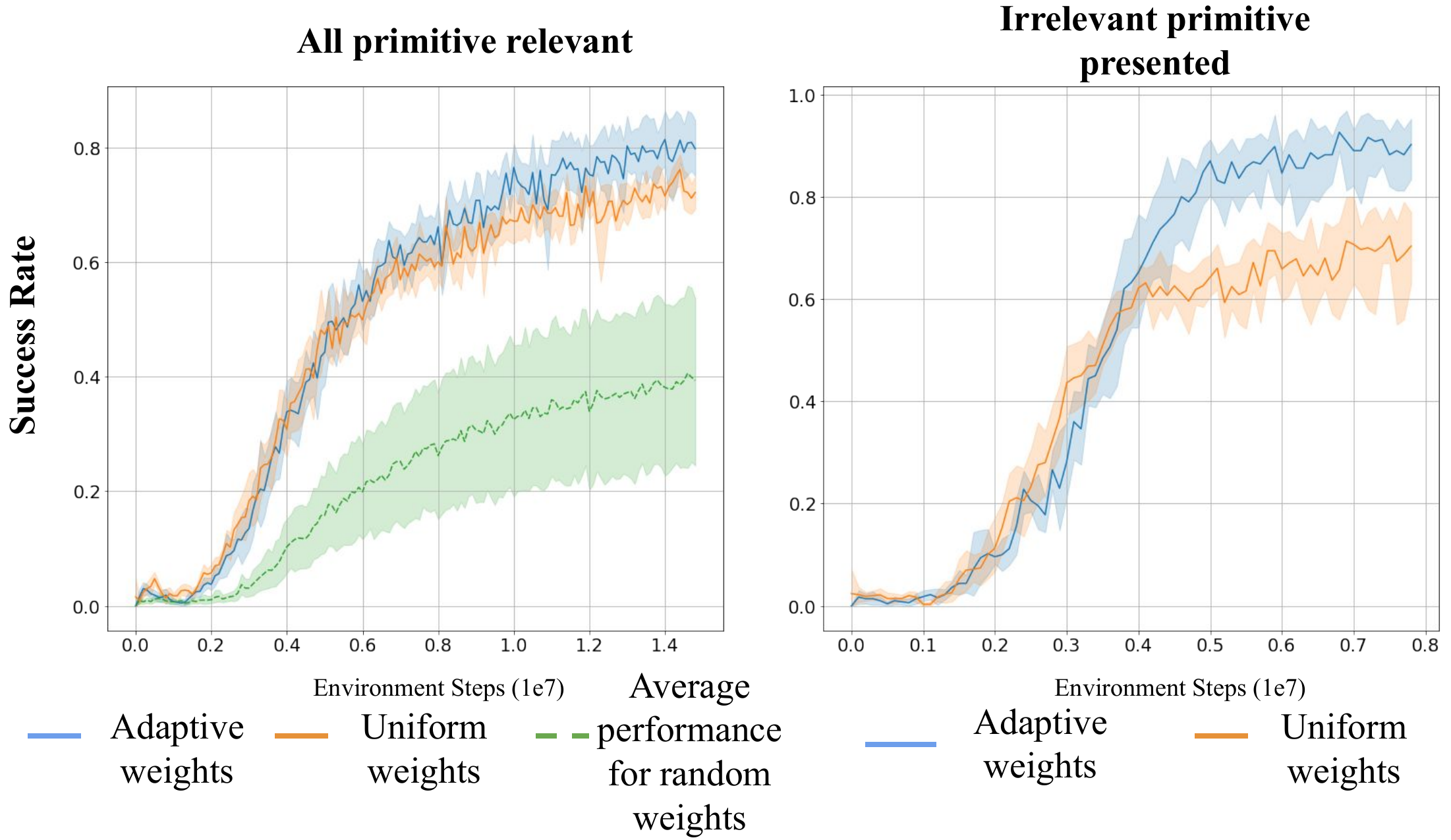}
\vspace{-3mm}
\caption{\textbf{Adaptive Weights vs Uniform Weights.} Learning curves of our method with Adaptive Weights and Uniform Weights in the setting (a) and (b). Notice that the tasks are different for this two settings. In the setting (a), The agent's goal is to traverse the maze while pushing the box to the goal position without hitting the obstacle. In the setting (b), the agent needs to reach the goal while avoiding the obstacle along the way without pushing the box.}
\label{fig:abl-uniform weights}
\end{center}
\vspace{-6mm}
\end{figure}
We have tested the uniform weight in the Ant Maze domain. Specifically, we conduct the experiment in two different settings: a) all skill priors primitives are relevant to the downstream tasks b) not all skill priors are relevant to the downstream tasks. 

In setting a) when all primitives are relevant, We observe that uniform weights also deliver good performance. If we compare it with many other random weights assignment (green line in Fig \ref{fig:abl-uniform weights}), uniform weights achieve a near-optimal assignment. The random weight performance is average of 6 different weight assignments. 
Regardless, our algorithm can still achieve the best performance among all weights assignments, showing that the system has enough flexibility to learn optimal policy. 

In the setting b) not all skill priors are relevant to the downstream tasks. Our experiment shows that the adaptive weights assignment significantly outperforms the uniform weights. This suggests that our adaptive weight assignment is robust to the various downstream tasks.

\subsection{Implementation Details}
\subsubsection{Offline phase}
The skill encoder is a one-layer LSTM with 128 hidden units, which receives $H=10$ steps action sequence as input and outputs the parameters of the Gaussian posterior $\mathcal{N}(\mu_z,\sigma_z)$ on a 10 dimensional skill embedding. The skill decoder is also a one-layer LSTM with 128 hidden units, which reconstructs the latent embedding back to the H steps action sequence, i.e., skill. Each primitive skill prior is parameterized with a 6-layer MLP with 128 hidden units per layer. We use ReLU as our activation layer and apply batch normalization. The network outputs parameters of the Gaussian skill prior $\mathcal{N}(\mu_a,\sigma_a)$. 

The models are optimized with Adam with learning rate \num{1e-4}and batch size 64. In the offline learning phase, we first learn a shared low-dimensional skill embedding space $\mathcal{Z}$ from aggregated datasets ${\{\mathcal{D}_i\}_{i=0}^K}$. For each iteration, we select one of the primitive datasets $\mathcal{D}_i$ and sample the state-skill tuples from it to optimize the skill embedding and the corresponding primitive skill prior $p_a^i(z|s)$. Once the ELBO loss in section 3.2 converges, we freeze the skill encoder/decoder parameters and only update the parameters of primitive skill prior. This makes sure that primitive skill priors can be learned on a stationary skill embedding space. We set the regularization parameter $\beta=$ \num{1e-4} for all experiment domains.
\subsubsection{Online phase}
 The learned policy $\pi_\theta(s)$ is parameterized with a 6-layer MLP with 128 hidden units per layer. The network outputs the parameters of a Gaussian distribution. We limit the action range of the policy between $[-2,...,2]$ by a tanh function. The weighting function $\omega_\sigma(s)$ is also parameterized with a 6-layer MLP with k-way softmax as the output layer. We use ReLU as our activation layer and apply batch normalization for both the learned policy and the weighting function. We model two critic networks and take the minimum value as Q-value estimation, which stabilizes the training. Each critic network is implemented as 4-layer MLP with 256 hidden units per layer. Skill prior generator $G_\sigma(z|s)$ is parameterized with a 3-layer MLP with 128 hidden units per layer. The network outputs the parameters of a Gaussian distribution.
 
 We optimized the models with Adam and set the learning rate as \num{1e-4}. The replay buffer capacity is 1e6, and the batch size is 256. We empirically find that setting the discount factor $\gamma$ as 0.97 can slightly stabilize the learning. The annealing coefficient $\beta$ starts with 1 and gradually decreases to \num{1e-3}. We set the target divergence $\delta=12$ for Point Maze, $\delta=15$ for Ant Push, and $\delta=20$ for Ant Maze. The temperature parameter $\alpha$ can be tuned automatically by minimizing the following:
\begin{align}
    \argmin_{\alpha>0} \mathbb{E}\big[\alpha\delta-\alpha\sum_{i=1}^K \omega_i(s_t)D_{KL}(\pi(z_t|s_t),p_a^i(z_t|s_t))\big]
\end{align}
The prove can be found in \citet{pertsch2020spirl}. We replace the single term KL divergence in the original formulation with weighted KL divergence. All models are trained on a single NVIDIA GPU.

\subsection{Environments and data collection}
For downstream tasks learning (online phase), the state $s_t$ can be partitioned into two components:
\begin{align}
    s_t = s_t^\text{p} + g_t \nonumber
\end{align}
where $s_t^\text{p}$ is the proprioceptive observation and $g_t$ is the downstream task-related observation. The states in each primitive dataset $\mathcal{D}_i$ always include the proprioceptive observation $s_t^\text{p}$, but part of the downstream task-related observation might be absent. For example, during the learning of Point Maze, the proprioceptive observation $s_t^\text{p}$ includes the agent's local view and velocity, and the task-related observation $g_t$ includes the positions of the obstacle, the goal, and the agent. However, in the navigation dataset, the whole task-related observation is absent. Suppose we extract the navigation primitive skill prior based on the proprioceptive observation only. In that case, the resulting prior cannot be directly applied to the downstream task as the state dimensions are different. To remedy this issue, we augment the states in offline data with random vectors that have the same length with the absent task-related observation. This augmentation will not impact the prior learning as the network will eventually ignore random vectors during the extracting process. Alternatively, instead of augmentation, we can parse the states during downstream task learning and only feed the relevant states to primitive skill priors. We have experienced both ways and find the choice does not impact the performance. For the experiment result in this paper, we apply augmentation. This issue might be solved by a method like Attention. However, this is out of the scope of this paper, and we will leave it for future research. 
\subsubsection{Point Maze}
The navigation dataset is collected by running the planning algorithm (provided in D4RL benchmark) in multiple random generated maze layouts. The avoid dataset is collected by a simple heuristic algorithm, i.e., the agent heads in any direction that can prevent itself from colliding with the obstacle. The heading direction $\theta$ can be calculated as: $\theta = -\frac{x_t^{agent}-x_t^{obstacle}}{||x_t^{agent}-x_t^{obstacle}||}\pm \gamma$, where $\gamma$ is a random number sampled from $[-\frac{\pi}{4},\frac{\pi}{4}]$, $x_t^{agent}$ is the position of the agent, and $x_t^{obstacle}$ is the position of the obstacle. The proprioceptive observation $s_t^\text{p}$ in primitive datasets includes the agent’s local view and  $v_t^{agent}$, which is the velocity of the agent.  

For the downstream task, the obstacles, goal, and agent positions are randomly reset at the beginning of one episode. We set a minimum distance between the goal and the agent position as 5 to avoid trivial tasks to constrain the reset process. The point agent will receive a sparse positive reward when the agent reaches the goal and a sparse negative reward when the agent hits the obstacle. The episode will be immediately reset once the agent receives either a positive or a negative reward. The additional task-related state $g_t$ is represented as a 30-dimensional vector which includes the local observation (i.e., goal/obstacle if in view) and the goal position. The action is a 2-dimensional continuous space that controls the agent's velocity in x and y-direction. 
\subsubsection{Ant Push}
To collect the push and avoid dataset, we first train a reaching policy by SAC, which allows the ant to head towards a specific target. The reaching policy takes a vector input $(x_t^{goal}-x_t^{agent},s_t^\text{p})$, where $x_t^{goal}$ is the goal position, $x_t^{agent}$ is the agent position, and proprioceptive observation $s_t^\text{p}$ is a 29-dimensional vector that represents the ant's joint state. The policy outputs an 8-dimensional vector to control the ant's joint. To collect the push dataset, we randomly place the ant at a distance of $[2,8]$ from the box. We use the reaching policy to create push behavior by setting the goal as the center of the box. The avoid dataset is generated by the same heuristic policy in the Point Maze environment.
\begin{figure}
    \centering
    \includegraphics[width=\textwidth]{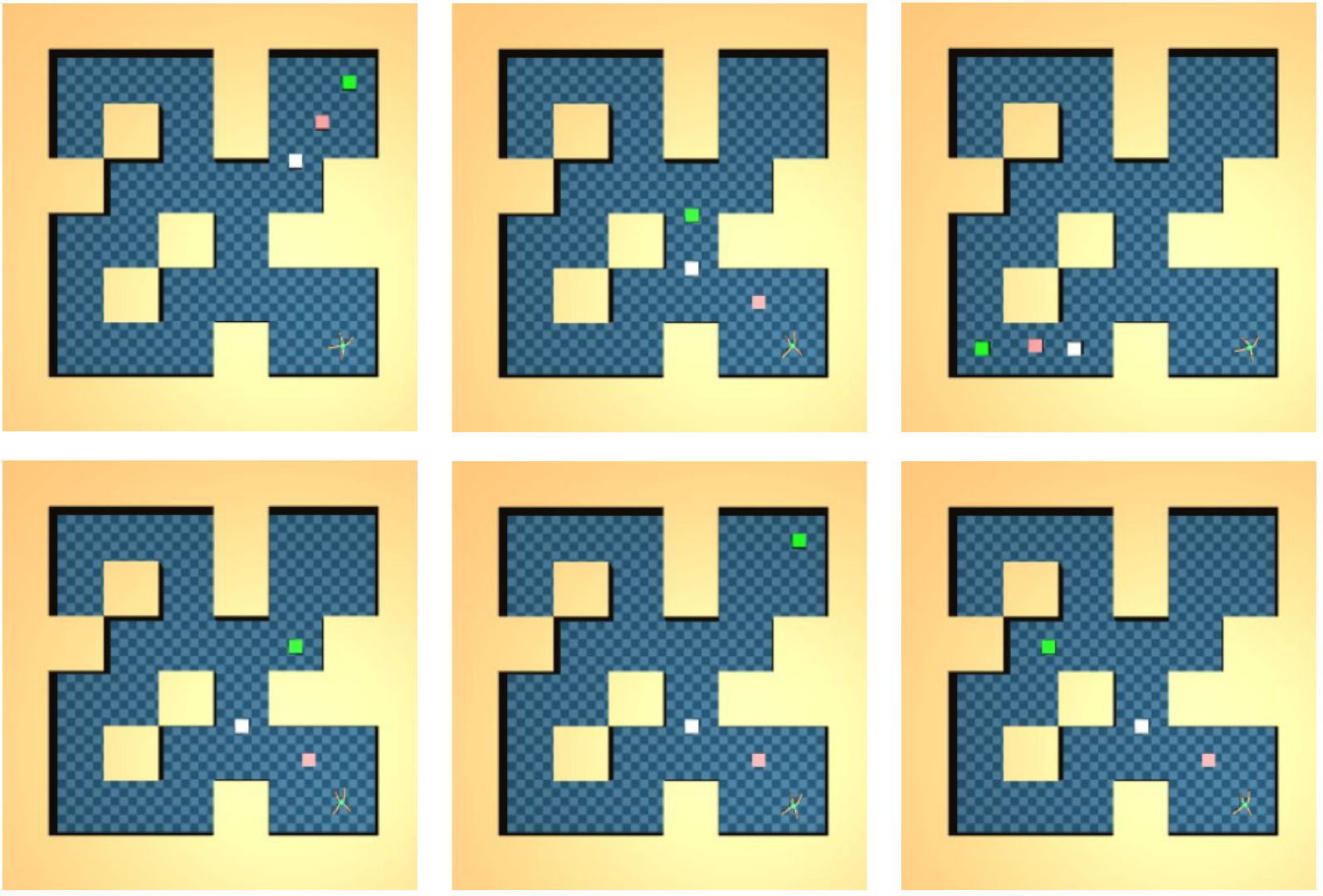}
    \caption{\textbf{Pre-defined sub-tasks in Ant Maze}}
    \label{fig:pre-defined tasks}
\end{figure}

For the downstream task, the obstacle, goal, and agent positions are randomly reset at the beginning of one episode. We randomly initialize the goal position $x_0^{goal}$ at a distance of $[-4,4]$ from the center $(0,0)$ and the agent position $x_0^{agent}$ at a distance of $[-8,8]$ from the center $(0,0)$. The obstacle is initialized between the goal and the agent $x_0^{obstacle}= \frac{x_0^{goal}+x_0^{agent}}{2}+ \epsilon$, where $\epsilon$ is sampled from Uniform(-1,1). Like Point Maze, we set a minimum initial distance between the goal and the agent position as 2 to avoid trivial cases. The agent will receive a sparse positive reward when the agent reaches the goal and a sparse negative reward when the agent hits the obstacle. The additional task-related state $g_t$ is a 4-dimensional vector representing the position difference between the agent and the goal $x_t^{goal}-x_t^{agent}$, and the position difference between the agent and the obstacle $x_t^{obstacle}-x_t^{agent}$. The action is an 8-dimensional continuous space. 

\subsubsection{Ant Maze}
We reuse the push and avoid dataset in Ant Push to extract the push and avoid primitive. Navigation primitive is extracted from the dataset (antmaze-medium-diverse) in D4RL benchmark. The proprioceptive observation $s_t^\text{p}$ is a 29-dimensional vector that represents the ant's joint state.

For the downstream task learning, we modify the Ant Maze environment from D4RL benchmark by adding an obstacle and a box in the maze. There are 6 pre-defined sub-tasks in the Ant Maze environment, and each of the sub-tasks specifies distinct obstacle, box, and goal positions. The agent will be given a task that is randomly selected from pre-defined sub-tasks at the beginning of every episode. The reward setting is the same as the one in Ant Push. However, the additional maze structure posts a hard exploration problem. The additional task-related state $g_t$ is a 6-dimensional vector representing the position difference between the agent and the goal $x_t^{goal}-x_t^{agent}$, and the position difference between the agent and the obstacle $x_t^{obstacle}-x_t^{agent}$ and the agent position $x_t^{agent}$. The action is an 8-dimensional continuous space. 

\subsubsection{Robotic Manipulation}
The task is to control a Panda robot arm to grasp the box without colliding with the barrier placed in the middle of the desk. The height of the barrier is set to be higher than the initial position of the robot arm's end effector. The location of the box and barrier are randomly generated at the beginning of the episode. The agent will receive a sparse positive reward once it grasps the box and lift it up without colliding with the barrier. ASPiRe will carry two primitive skill priors: grasp and avoid. The grasp data is generated by a hand-coded heuristic policy, which first moves the end effector horizontally to the top of the object and then moves the end effector down to grasp the object. The avoid data is also generated by a hand-coded heuristic policy, which moves the end effector up vertically once the barrier is observed. As the datasets are collected by heuristic policies, a part of trajectories might fail to complete the primitive tasks, i.e., grasp the box. Therefore, we only keep the trajectories that successfully complete the primitive tasks.

\end{document}